\documentclass[fleqn,10pt,twocolumn]{wlscirep}
\usepackage[utf8]{inputenc}
\usepackage[T1]{fontenc}

\usepackage{graphicx,float}
\usepackage{subfig}
\usepackage{cite}
\usepackage{multirow}
\usepackage{amsmath,amssymb,amsfonts}
\usepackage{algorithmic}
\usepackage{lmodern}
\usepackage{titlesec}
\usepackage{times}
\usepackage[symbol]{footmisc}
\usepackage{hyperref}
\usepackage{lettrine}
\usepackage{xcolor}
\usepackage[utf8]{inputenc}
\DeclareUnicodeCharacter{FB01}{fi}

\usepackage{authblk}
\usepackage{hyperref}

\newbox{\myorcidaffilbox}
\sbox{\myorcidaffilbox}{\large\includegraphics[height=4mm]{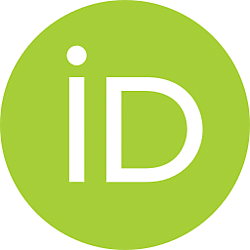}}
\newcommand{\orcidaffil}[1]{\href{https://orcid.org/#1}{\usebox{\myorcidaffilbox}}}
 
\newbox{\mybox}
\sbox{\mybox}{\large\includegraphics[height=2.5mm]{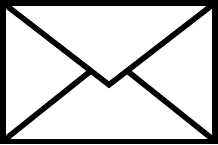}}
\newcommand{\boxaffil}[1]{\href{https://orcid.org/#1}{\usebox{\mybox}}}

\usepackage{lipsum,microtype}
\usepackage{subfig}
\usepackage{framed,multirow}
\usepackage[symbol]{footmisc}
\usepackage{amssymb}
\usepackage{latexsym}

\usepackage{url}
\usepackage{xcolor}
\usepackage{hyperref}
\definecolor{newcolor}{rgb}{.8,.349,.1}

\usepackage{geometry}
\usepackage{fleqn}
\usepackage{newtxtext,newtxmath}
\usepackage{amsfonts}
\usepackage{scalerel}
\usepackage[edges]{forest}
\forestset{declare boolean={switch me}{0}}
\usepackage{graphicx,float}
\usepackage[export]{adjustbox}
\usepackage{graphicx}
\DeclareMathAlphabet{\mathpzc}{OT1}{pzc}{m}{it}
\usepackage{wrapfig,booktabs}
\usepackage{multirow,  makecell, booktabs, tabularx}
\usepackage{tabularx,ragged2e}
\usepackage[export]{adjustbox}
\usepackage{amsmath}
\usetikzlibrary{arrows.meta}

\usepackage{hyperref}
\hypersetup{
    colorlinks=true,
    linkcolor=red,
    filecolor=magenta,      
    urlcolor=cyan,
}
\usepackage{blindtext}
\usepackage{etoolbox}
\usepackage{pgfplots}
\usepackage{pgfplotstable}
\pgfplotsset{compat=1.8,
        /pgfplots/ybar legend/.style={
        /pgfplots/legend image code/.code={%
        \draw[##1,/tikz/.cd,bar width=3pt,yshift=-0.2em,bar shift=0pt]
                plot coordinates {(0cm,0.8em)};},
},
}
\usepackage[utf8]{inputenc}
\usetikzlibrary{trees}
\DeclareMathOperator*{\argmax}{argmax} 
\tikzset{font={\fontsize{5.5pt}{8}\sffamily}}
\definecolor{electriclime}{rgb}{0.85, 1.0, 0.0}
\definecolor{yellow}{rgb}{1.0, 1.0, 0.0}
\definecolor{azurecw}{rgb}{0.0, 0.7, 1.0}
\definecolor{amber}{rgb}{1.0, 0.75, 0.0}
\definecolor{bostonuniversityred}{rgb}{0.8, 0.0, 0.0}
\definecolor{tractorred}{rgb}{1.0, 0.0, 0.25}
\definecolor{cadmiumred}{rgb}{0.89, 0.0, 0.13}
\definecolor{ao}{rgb}{0.0, 0.0, 1.0}
\definecolor{safetyorange}{rgb}{1.0, 0.4, 0.0}
\definecolor{spirodiscoball}{rgb}{0.21, 0.44, 0.98}
\definecolor{aqua}{rgb}{0.0, 1.0, 1.0}
\definecolor{guppiegreen}{rgb}{0.0, 1.0, 0.5}
\definecolor{ufo}{rgb}{0.55, 0.98, 0.0}
\definecolor{darkmagneta}{rgb}{0.55, 0.0, 0.55}
\definecolor{columbiablue}{rgb}{0.61, 0.87, 1.0}
\definecolor{hotpink}{rgb}{1.0, 0.41, 0.71}
\definecolor{brightube}{rgb}{0.82, 0.62, 0.91}
\makeatletter

\patchcmd\@footnotetext
  {\hsize\columnwidth}
  {\hsize\columnwidth\advance\hsize\marginparsep\advance\hsize\marginparwidth}
  {}{}

\pgfkeys{/pgf/.cd,
 parallelepiped offset x/.initial=2mm,
 parallelepiped offset y/.initial=5mm
}
\pgfdeclareshape{parallelepiped}
{
  \inheritsavedanchors[from=rectangle] 
  \inheritanchorborder[from=rectangle]
  \inheritanchor[from=rectangle]{north}
  \inheritanchor[from=rectangle]{north west}
  \inheritanchor[from=rectangle]{north east}
  \inheritanchor[from=rectangle]{center}
  \inheritanchor[from=rectangle]{west}
  \inheritanchor[from=rectangle]{east}
  \inheritanchor[from=rectangle]{mid}
  \inheritanchor[from=rectangle]{mid west}
  \inheritanchor[from=rectangle]{mid east}
  \inheritanchor[from=rectangle]{base}
  \inheritanchor[from=rectangle]{base west}
  \inheritanchor[from=rectangle]{base east}
  \inheritanchor[from=rectangle]{south}
  \inheritanchor[from=rectangle]{south west}
  \inheritanchor[from=rectangle]{south east}
  \backgroundpath{
    \southwest \pgf@xa=\pgf@x \pgf@ya=\pgf@y
    \northeast \pgf@xb=\pgf@x \pgf@yb=\pgf@y
    \pgfmathsetlength\pgfutil@tempdima{\pgfkeysvalueof{/pgf/parallelepiped offset x}}
    \pgfmathsetlength\pgfutil@tempdimb{\pgfkeysvalueof{/pgf/parallelepiped offset y}}
    \def\ppd@offset{\pgfpoint{\pgfutil@tempdima}{\pgfutil@tempdimb}}
    \pgfpathmoveto{\pgfqpoint{\pgf@xa}{\pgf@ya}}
    \pgfpathlineto{\pgfqpoint{\pgf@xb}{\pgf@ya}}
    \pgfpathlineto{\pgfqpoint{\pgf@xb}{\pgf@yb}}
    \pgfpathlineto{\pgfqpoint{\pgf@xa}{\pgf@yb}}
    \pgfpathclose
    \pgfpathmoveto{\pgfqpoint{\pgf@xb}{\pgf@ya}}
    \pgfpathlineto{\pgfpointadd{\pgfpoint{\pgf@xb}{\pgf@ya}}{\ppd@offset}}
    \pgfpathlineto{\pgfpointadd{\pgfpoint{\pgf@xb}{\pgf@yb}}{\ppd@offset}}
    \pgfpathlineto{\pgfpointadd{\pgfpoint{\pgf@xa}{\pgf@yb}}{\ppd@offset}}
    \pgfpathlineto{\pgfqpoint{\pgf@xa}{\pgf@yb}}
    \pgfpathmoveto{\pgfqpoint{\pgf@xb}{\pgf@yb}}
    \pgfpathlineto{\pgfpointadd{\pgfpoint{\pgf@xb}{\pgf@yb}}{\ppd@offset}}
  }
}

\title{Zero-Shot Learning and its Applications from Autonomous Vehicles to COVID-19 Diagnosis: A Review}
\date{}
\author{
Mahdi Rezaei \orcidaffil{0000-0003-3892-421X}  $^{1, \, \boxaffil{}}$ , \, Mahsa Shahidi \orcidaffil{0000-0003-3724-4597}  $^{2, \star}$ \\ 
$^1$ Institute for Transport Studies, The University of Leeds, Leeds, LS2 9JT, UK\\ 
$^2$ Department of Computer Engineering, Qazvin Azad University, Qazvin, IR\\ 
$^1$ \href{mailto:m.rezaei@leeds.ac.uk}{m.rezaei@leeds.ac.uk} \,\, $^2$ \href{mailto:m.shahidi@qiau.ac.ir}{m.shahidi@qiau.ac.ir}
}

\begin{abstract}
The challenge of learning a new concept, object, or a new medical disease recognition without receiving any examples beforehand is called Zero-Shot Learning (ZSL). One of the major issues in deep learning based methodologies such as in Medical Imaging and other real-world applications is the requirement of large annotated datasets prepared by clinicians or experts to train the model. ZSL is known for having minimal human intervention by relying only on previously known or trained concepts plus currently existing auxiliary information. This is ever-growing research for the cases where we have very limited or no annotated datasets available and the detection$\slash$recognition system has human-like characteristics in learning new concepts. This makes the ZSL applicable in many real-world scenarios, from unknown object detection in autonomous vehicles to medical imaging and unforeseen diseases such as COVID-19 Chest X-Ray (CXR) based diagnosis. In this review paper, we introduce a novel and broaden solution called Few$\slash$one-shot learning, and present the definition of the ZSL problem as an extreme case of the few-shot learning. We review over fundamentals and the challenging steps of Zero-Shot Learning, including state-of-the-art categories of solutions, as well as our recommended solution, motivations behind each approach, their advantages over each category to guide both clinicians and AI researchers to proceed with the best techniques and practices based on their applications. Inspired from different settings and extensions, we then review through different datasets inducing medical and non-medical images, the variety of splits, and the evaluation protocols proposed so far. Finally, we discuss the recent applications and future directions of ZSL. We aim to convey a useful intuition through this paper towards the goal of handling complex learning tasks more similar to the way humans learn. We mainly focus on two applications in the current modern yet challenging era: coping with an early and fast diagnosis of COVID-19 cases, and also encouraging the readers to develop other similar AI-based automated detection$\slash$recognition systems using ZSL.\\


\end{abstract}

\begin{document}
\flushbottom
\maketitle
\thispagestyle{empty}

\footnote[0]{$^{\includegraphics[width=3mm]{drawing.pdf} \href{https://orcid.org/0000-0003-3892-421X}{}}$ Corresponding Author: \href{mailto:m.rezaei@leeds.ac.uk}{m.rezaei@leeds.ac.uk} (M. Rezaei)}

\vspace{-10mm}
\section{Introduction}\label{sec:introduction}

\lettrine[lines=3]{\textcolor[rgb]{0.4,0.4,0.4}O}{\, bject} 
recognition is one of the highly researched areas of computer vision. Recent recognition models have led to great performance through established techniques and large annotated datasets. After several years of research, the attention over this topic has not only dimmed but it has been proven that there are still ways and rooms to refine models to eliminate existing issues in this area. 
The number of newly emerging unknown objects are growing. Some examples of these unseen or rarely-seen objects are 
futuristic object designs like the next generation of concept cars, other existing concepts but with restricted access to them (such as licensed or private medical imaging datasets), or rarely seen objects (such a traffic signs with graffiti on them), or fine-grained categories of objects (such as detection of COVID-19 in comparison with the easier task of detecting a common pneumonia). This brings the necessity of developing a fresh way of solving object recognition problems that concern lesser human supervision and lesser annotated datasets. 
Several approaches have tried to gather web images to train the developed deep learning models, but aside from the problem of the noisy images, the searched keywords are still a form of human supervision. One-Shot learning (OSL) and Few-shot learning (FSL) are two solutions that are able to learn new categories via one or a few images, respectively \cite{miller2000learning}, \cite{lake2011one}, \cite{koch2015siamese}. 

\begin{figure*}
\centering
\subfloat[Positive COVID-19]
{\resizebox*{0.66\linewidth}{!}{\includegraphics{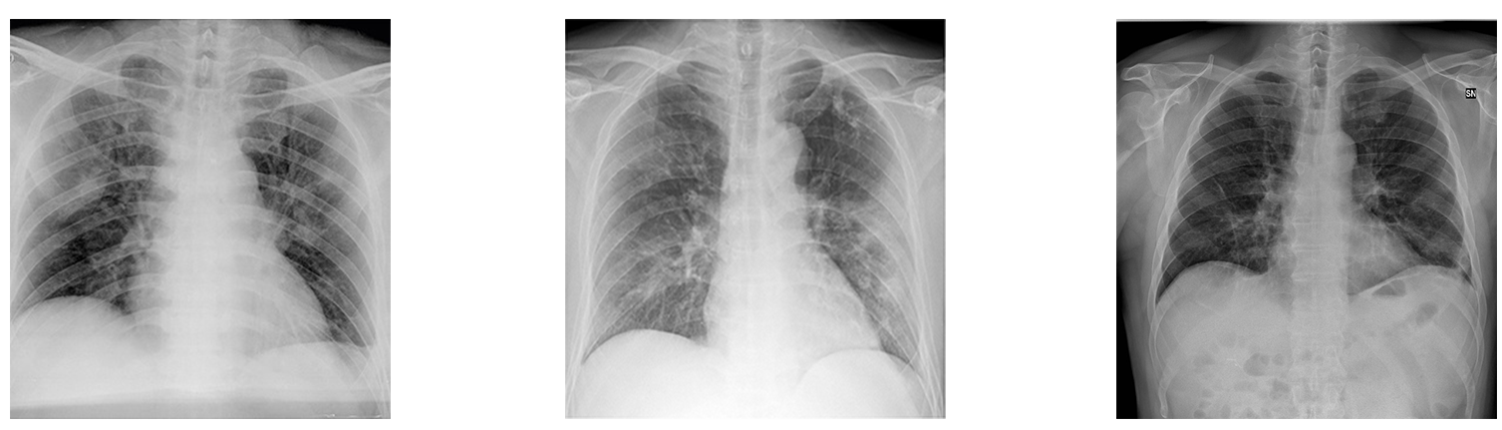}\label{a}}}\\
\subfloat[Axial chest CTs with visible GGO patches, consolidation, and parenchymal thickening]
{\resizebox*{0.77\linewidth}{!}{\includegraphics{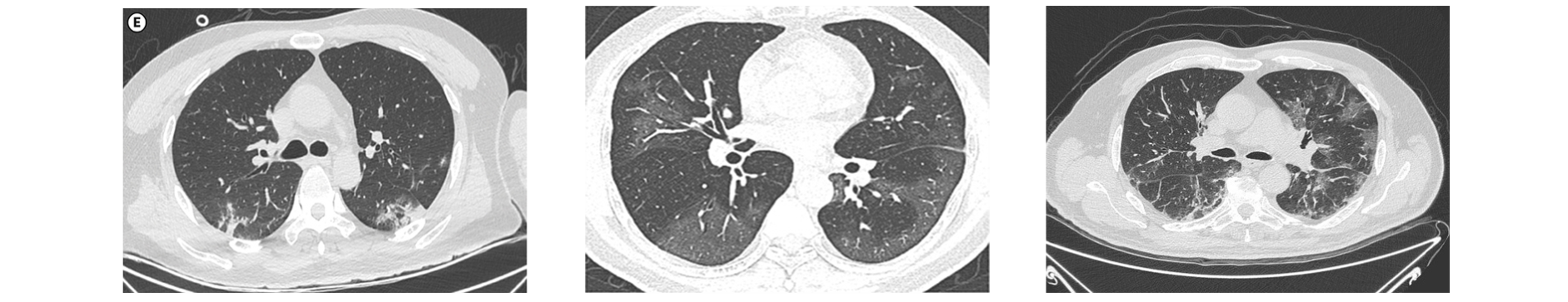}\label{b}}}
\caption{Posterior-Anterior (PA)/ Anterior-Posterior (AP) Chest X-Rays and the corresponding CT images of COVID-19 patients}
\vspace{-1mm}
\label{CXR}
\end{figure*}

Natural language processing (NLP) is another major area of research in AI and the application of Few-shot learning in the integration of NLP and object recognition has become a hot topic recently. \cite{tsai2017improving} was the first FSL-based model to improve the performance of an NLP system. Zero-shot learning (ZSL) \cite{lampert2013attribute}, \cite{akata2015evaluation}, \cite{xian2017zero}, \cite{elhoseiny2017link}, \cite{wang2019survey} is an emerging research which is completely free of any laborious task of data collection and annotation by experts. Zero-shot learning is a novel concept 
and learning technique without accessing any exemplars of the unseen categories during training, yet it is able to build recognition models with the help of transferring knowledge from previously seen categories and auxiliary information. The auxiliary information may include textual description, attributes, or vectors of word labels. This means the ZSL is interdisciplinary by nature with two inseparable components of visual and textual data.

One of the interesting facts about ZSL is its similarity with the way human learns and recognise a new concept without seeing them beforehand. For example, a ZSL-based model would be able to automatically learn and diagnose COVID-19 patients, based on the existing chest X-ray images of patients with asthma and lungs inflammatory diseases which are already recognised and labelled by clinicians, plus some new auxiliary information about the COVID-19 attributes. Here, the auxiliary data can be the description of physicians and clinicians about the unique type of visual patterns, features, damages, or differences they have noticed on the Chest X-ray of patients with positive COVID-19 comparing to asthma X-ray images.
A similar concept or approach is applicable in autonomous vehicles, \cite{Rezaei2014a}, where a self-driving car is responsible for automatic detection of surrounding cars including e.g. an unseen Tesla concept car based on the subgroup of labelled classic sedan cars plus auxiliary information about the common differences of concept cars than the classic cars; or recognising a Persian deer, based on the auxiliary information available for it and its appearance similarities or differences with other previously known deer. For instance, it belongs to a subgroup of the fallow deer, but with a larger body, bigger antlers, white spots around the neck, and also flat antlers for the male type.

\begin{figure*}[t!]
\centering
\subfloat[Concept cars auxiliary information: \textit{``The body of the car has a singular and unified shape with smoother curves. The wheels' colour, curves, and design match the body as a singular integrated piece. LED lights are omnipresent all around the car.''}]{
\resizebox*{0.505\linewidth}{!}{\includegraphics{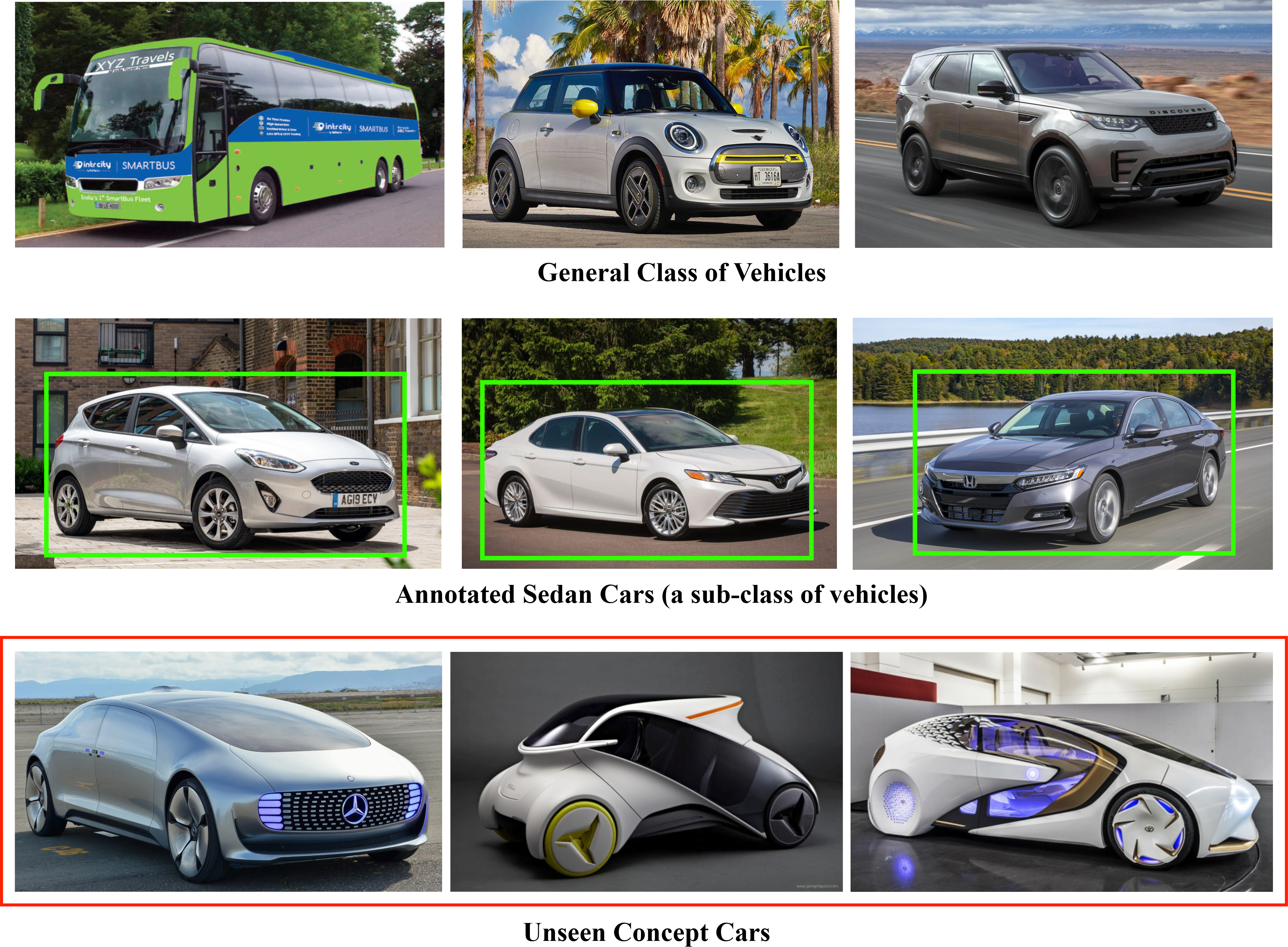}\label{conceptcars}}}\hspace{25pt}
\subfloat[COVID-19 X-ray auxiliary information: \textit{``Bilateral multifocal patchy GGOs and consolidation can be seen. Edges are blurred and the intensity sharpness of both lungs have decreased.''}]{
\resizebox*{0.41\linewidth}{!}{\includegraphics{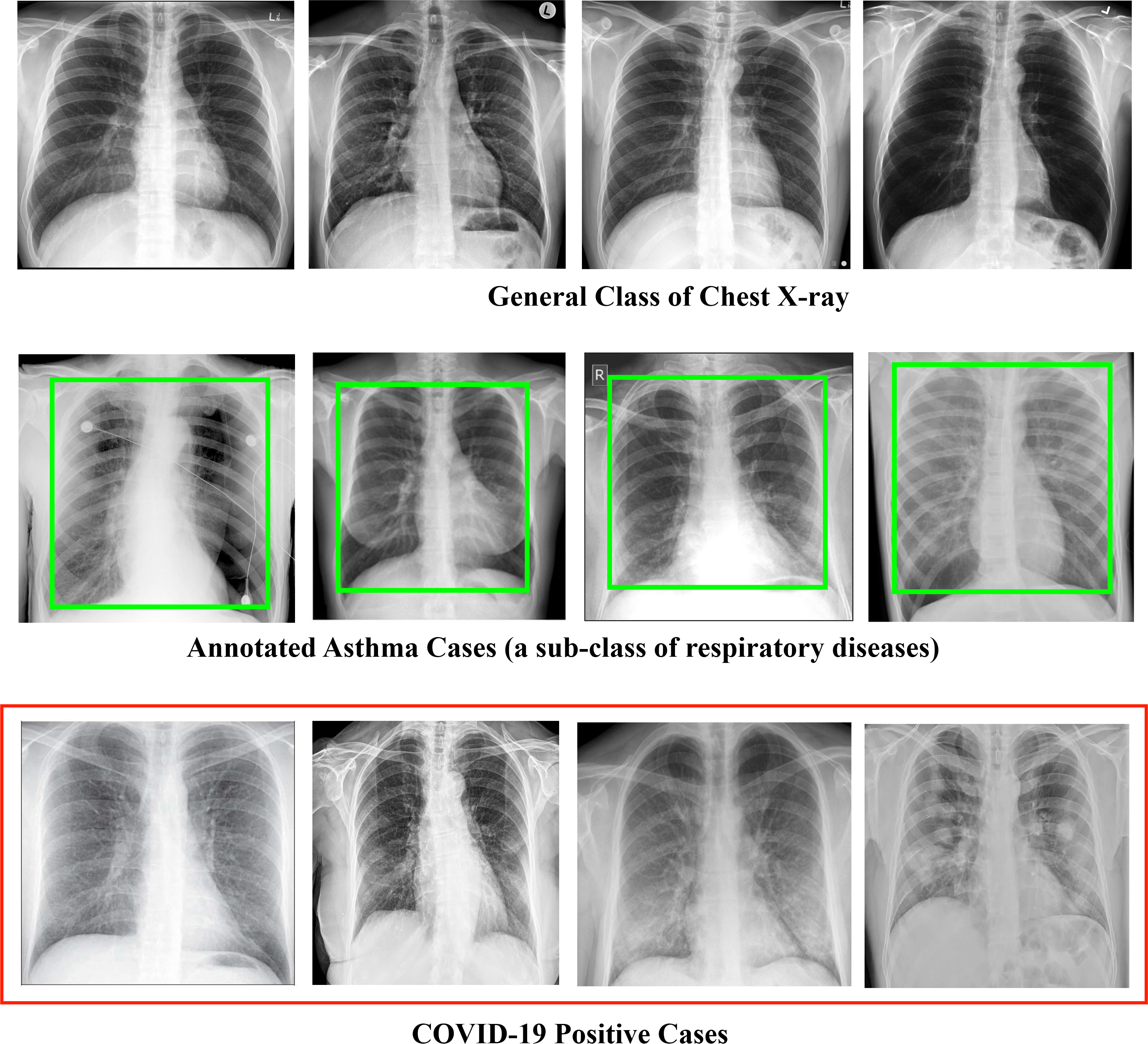}\label{covidxrays}}}
\caption{Similarities and differences between seen and unseen examples derived from textual descriptions and train and test images. The test images are concept cars (a) and COVID-19 symptoms (b).}
\label{auxiliary-fig}
\end{figure*}

Figure \ref{CXR}\subref{a} shows three examples of Posterior-Anterior (PA) and AP projection of chest X-rays of positive cases of COVID-19, and Figure \ref{CXR}\subref{b} represents their corresponding axial CT scans, taken from the COVID-ChestXRay dataset \cite{cohen2020covid}. As it can be seen in the images, common evident anomalies may include unilateral or bilateral patchy ground-glass opacities (GGOs), patchy consolidations and parenchymal thickening. 
The goal of this research is to build an artificial intelligence based-model that can diagnoses COVID-19 without providing any visual exemplars in the training phase. In that case, the side (auxiliary) information should be provided to assist diagnosis in the test phase. In Figure \ref{auxiliary-fig}, the auxiliary information is provided in the form of textual descriptions for two examples of concept cars and COVID-19 X-rays. In Figure 
\ref{auxiliary-fig}\subref{conceptcars} we aim at distinguishing new unseen concept cars (bottom row), using description on the exterior of the target and how it differs an already learned car from existing classic vehicle classification system such as in \cite{rezaei2015}. Similarly, visual differences and similarities between healthy Chest X-rays, Asthma cases, and COVID-19 positive cases are described in Figure \ref{auxiliary-fig}\subref{covidxrays} as the auxiliary information. 

Let's assume our pre-trained AI-based medical imaging system is capable of detecting Asthma cases, based on common deep learning techniques using a previously large dataset of labelled Asthma Chest X-ray images. However, these days we are facing an unknown COVID-19 pandemic with very limited annotated Chest X-rays. Obviously, we can not proceed on the same way of training traditional deep-learning methods, due to very sparse labelled images for COVID-19. The good point is that our medical experts and clinicians can provide some auxiliary information (textual descriptions) about common features and similarities among the COVID-19 positive chest X-rays to infer their findings. In Figure \ref{fig:overview}, the side information is provided in form of what \textit{``attributes''}: such as foggy effects, white spot features, blurred edges, and white/low-intensity pixel dominance in various areas of the chest X-ray images of COVID-19 patients.\\

Our idea behind the utilisation of ZSL models is to detect, understand, and recognise new concepts using an existing similar deep-learning based classifier, plus the integration of auxiliary information. This turns it to a completely new and efficient detector/recogniser or diagnosing system without the requirement of collecting a new dataset and a vast amount of costly and time-consuming labelling, especially when a speedy solution is crucial and life-saving; such as the recent global pandemic.\\

\noindent In this research we will have four main contributions, as follows: 
\begin{itemize}
\item We propose to categorise the reviewed approaches based on the embedding spaces that each model uses to learn/infer unseen objects/concepts as well as describing the variations to the data embedding inside those embedding spaces
(Figure~\ref{fig:overview} and Table \ref{tab:summary}). 
\vspace{-1mm}
\item We evaluate the performance of the state-of-the-art models on famous benchmark datasets (Table \ref{tab:ssps}--\ref{tab:scssce}, Fig. \ref{fig4}). 
To the best of our knowledge, we are the first to include the evaluation of data-synthesising methods in the research field of applied Zero-shot learning. 
\vspace{-1mm}
\item We study the motivation behind leveraging each space as a way to solve the ZSL challenge by reviewing current issues and solutions to them.
\vspace{-1mm} 
\item We provide sufficient technical justifications to support the ideas of using the proposed ZSL model as one of the best practices for COVID-19 diagnosis and other similar applications. 
\end{itemize}

The rest of the materials in the article is organised as follows. In Section \ref{zero-few}, we introduce the problem of Few-shot, One-shot and Zero-shot learning. In Section \ref{ZSL_TT}, we discuss about the test and train phases of the Zero-shot learning and generalised Zero-shot learning systems. Section \ref{appro} provides with embedding approaches followed by evaluation protocols in Section \ref{eval}. In Section \ref{exper}, we analyse the outcome of the experiments performed on different state-of-the-art methodologies. Further discussion about the applications of ZSL is investigated in Section \ref{App}. In Section \ref{discuss}, we discuss the outcome of this research, and finally, the concluding remarks in Section \ref{conc}.\\

\vspace{-4mm}
\section{Few-Shot / One-Shot and Zero-Shot Learning \label{zero-few}}
\noindent Few-shot learning (FSL) is the challenge of learning novel classes with a tiny training dataset of one or a few images per category. FSL is closely related to knowledge transfer where a model, previously trained on large data, is used for a similar task with fewer training data. The more the transferred knowledge is accurate, the better FSL will generalise. Moreover, many approaches employ meta-learning to learn the challenge of few-shot or few-example learning \cite{snell2017prototypical}, \cite{kang2019few}. The main challenge is to improve the generalisation ability as it often faces the overfitting problem.

In this type of learning, there is an auxiliary dataset that contains $N$ classes each having $K$ annotated samples of the new examples in the training phase. This makes the problem into a N-way-K-shot classification:
\begin{align}
D_{s}=\{(x_i,y_i)\}^{N_{t}}_{i=1}
\end{align}

\noindent where $x_i$ is the $i^{th}$ training example and $y_i$ is its corresponding label. $N_{t}=K\times N$ denotes the number of N categories and $K$ defines the number of examples. Few-shot learning has $K$ > 1 samples. 

Among the relevant research works, \cite{torralba2006shared} use the shared features among classes to compensate for the requirement for the large data, and follows a learning procedure based on boosted decision stumps. HDP-DBM \cite{salakhutdinov2012learning} develops a compound of a deep Boltzmann machine and a hierarchical Dirichlet process to learn the abstract knowledge at different hierarchies of the concept categories. \cite{snell2017prototypical} Proposes prototypical networks that computes Euclidean distance between prototype representations of each class. 
It was not until recently that Few-shot learning was introduced in computer-aided diagnosis. 
For the first time, the idea of using additional information (attributes) in FSL, was introduced in \cite{tsai2017improving}.
\cite{prabhu2019few} proposes a model to classify skin lesions. \cite{kim2017few} use FSL for Glaucoma Diagnosis from fundus images. \cite{rajan2020self} study the problem of chest X-ray classification of five symptoms including Consolidation.  

In the case of one-shot learning, there is only $K=1$ example per class in the supporting set, thus it faces more challenge in comparison to the FSL. Bayesian Program Learning (BPL) framework \cite{lake2015human} present each concept of the handwritten characters as a simple probabilistic program. \cite{bart2005cross} proposes cross-generalisation algorithm. It replaces the features from the previously learned classes with similar features of the novel classes to adapt to the target task. In Bayesian learning, \cite{fei2006one} depicts prior knowledge in the form of probability density function on the parameters of the model, and updates them to compute the posterior model. Matching Nets (MN) \cite{vinyals2016matching} uses non-parametric attentional memory mechanisms, and an \textit{``episode''} during the training time. \cite{chen2020momentum} capture salient features of general lung datasets using an encoder and augment multiple views for images, then use the prototypical network for a 2-way, 1-shot classification. 

Zero-shot learning is the extreme case of the FSL where $K=0$. In other words, the difference between the two is the devoid of any visual examples of the target classes in the training phase of ZSL, while in few-shot learning, the support set contains few labelled samples of the novel categories. Also, auxiliary information in the form of class embeddings is one of the main components of Zero-shot learning. ZSL approaches might extend their solutions to one-shot or few-shot learning by either updating the training data with one or few generated samples from augmentation techniques, or by having access to a few of the unseen images during the training time \cite{schonfeld2019generalized}, \cite{yu2010attribute}, \cite{sharmanska2012augmented}, \cite{akata2013label}, \cite{jayaraman2014zero}, \cite{bucher2016improving}, \cite{verma2017simple} \cite{changpinyo2017predicting}, \cite{tsai2017learning}, \cite{xian2019f}, \cite{schonfeld2019generalized}. \cite{xian2019f} and \cite{schonfeld2019generalized} both use auxiliary text-based information. 

\vspace{-2mm}
\section{ZSL Test and Training Phases}
\label{ZSL_TT}
\vspace{-0.1cm}
\noindent ZSL models can be seen from two points of views in terms of training and test phase: Classic ZSL and Generalised ZSL (GZSL) settings. In the classic ZSL settings, the model only detects the presence of new classes at the test phase, while in GZSL settings, the model predicts both unseen and seen classes at the test time; hence, GZSL is more applicable for real-world scenarios \cite{liu2018generalized}, \cite{kumar2018generalized}, \cite{zhu2019generalized}, \cite{li2019generalized}, \cite{schonfeld2019generalized}. The same idea can be applied to FSL to  train in the generalised model, called generalised few-shot learning (GFSL) that detects both known and novel classes at the test time.

In the next paragraphs, we discuss two types of training approaches: Inductive vs. Transductive training.\\

\noindent \textbf{Inductive Training}: This training setting only uses the seen class of information to learn a new concept. The training data for the inductive setting is:
\begin{equation}
D_t=\{(x,y,c(y))|x\in X^S,y\in Y^S, c(y) \in C^S\}	
\end{equation}

\noindent where $x$ represents image features, $y$ is the class labels, and $c(y)$ denotes the class embeddings. Moreover, $X^S$ and $Y^S$ indicate seen class images and seen class labels, respectively. Inductive learning accounts for the majority of the settings used in ZSL and Generalised Zero-Shot Learning (GZSL). e.g. in~\cite{akata2015evaluation}, \cite{frome2013devise}, \cite{norouzi2013zero}, \cite{romera2015embarrassingly}, \cite{guo2017zero}, \cite{zhang2015zero}, \cite{changpinyo2016synthesized}, \cite{zhao2017zero}, \cite{li2017zero}, \cite{verma2017simple}, \cite{xian2019f}.\\

\noindent \textbf{Transductive Training}:
Although the original idea of zero-shot learning is more related to the inductive setting, in many scenarios, the transductive setting is used where either unlabelled visual or textual information, or both for unseen classes are used together with the seen class data e.g. in in~\cite{rohrbach2013transfer}, \cite{kodirov2015unsupervised}, \cite{fu2015transductive}, \cite{akata2015label}, \cite{zhang2016zero}, \cite{xu2017transductive}, \cite{guo2017zero}, \cite{zhao2017zero}, \cite{li2017zero}, \cite{verma2017simple}, \cite{song2018transductive}, \cite{wan2019transductive}, \cite{xian2019f}, \cite{sariyildiz2019gradient}. The training data for transductive learning is:
\begin{equation}
D_t=\{(x,y,c(y))|x\in X^{S\cup U},y\in Y^{S\cup U}, c(y) \in C^{S\cup U}\}
\end{equation}

\noindent where $X^{S\cup U}$ denotes that images come from the union of seen and unseen classes. Similarly, $Y^{S\cup U}$ and $C^{S\cup U}$ indicate the train labels and class embeddings belong to both seen and novel categories.

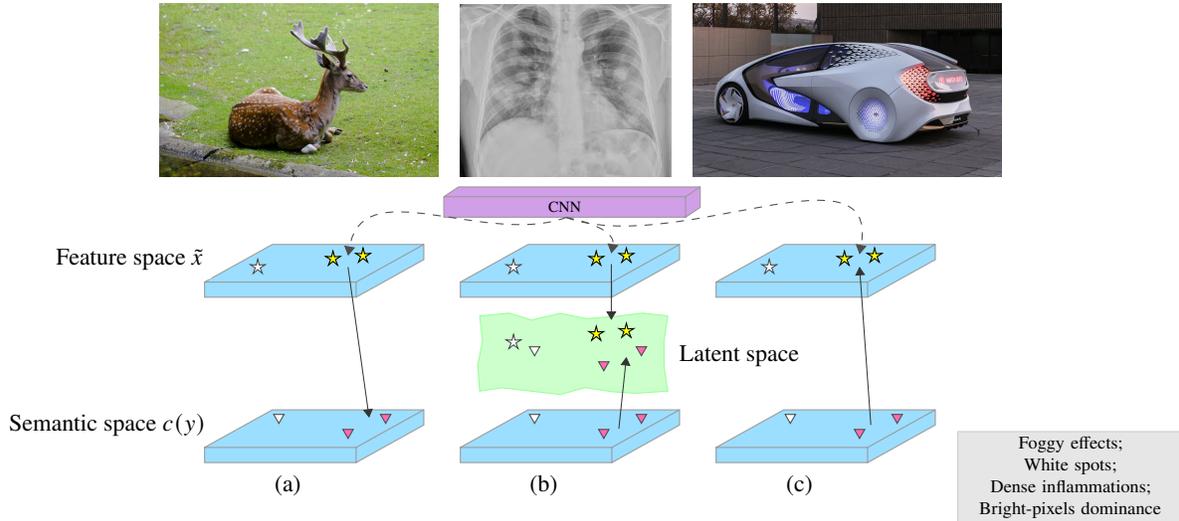
\begin{figure*}
\vspace{1.7cm}
\centering
\footnotesize
\begin{tikzpicture}
\tikzstyle{every node}=[font=\small,align=center]
\node[parallelepiped,draw=black!40,fill=columbiablue,
  minimum width=2cm,minimum height=0.01cm,parallelepiped offset x=9mm] (1) at (6.5,-1.5){};
\node (rect) at (4.5,-1.1) [draw=white,minimum width=2cm,minimum height=0.5cm] {Feature space $\tilde {x}$};
\node (rect) at (4.2,-3.3) [draw=white,minimum width=2cm,minimum height=0.5cm] {Semantic space $c(y)$};
\node (rect) at (12.6,-2.38) [draw=white,minimum width=2cm,minimum height=0.6cm] {Latent space};
\node (rect) at (6.6,-4.1) [draw=white,minimum width=2cm,minimum height=0.6cm] {(a)};
\node (rect) at (10,-4.1) [draw=white,minimum width=2cm,minimum height=0.6cm] {(b)};
\node (rect) at (13.4,-4.1) [draw=white,minimum width=2cm,minimum height=0.6cm] {(c)};
\node [fill=white,scale=0.26,star,star point ratio= 3,draw=black!70] (9) at (6.2,-1.2) {};
\node [fill=yellow,scale=0.26,star,star point ratio= 3,draw] (9) at (7.6,-1.05) {};
\node [fill=yellow,scale=0.26,star,star point ratio= 3,draw] (15) at (7.2,-1.09) {};
\node[parallelepiped,draw=black!40,fill=columbiablue,
  minimum width=2cm,minimum height=0.01cm,parallelepiped offset x=9mm] (2) at (6.5,-3.7){};
\node [fill=white,scale=0.3,regular polygon,regular polygon sides=3, draw=black!70, shape border rotate=180] (16) at (6.48,-3.2) {};
\node [fill=hotpink,scale=0.3,regular polygon,regular polygon sides=3, draw=black!70, shape border rotate=180] (17) at (7.4,-3.4) {};
\node [fill=hotpink,scale=0.3,regular polygon,regular polygon sides=3, draw=black!70, shape border rotate=180] (18) at (7.9,-3.2) {};
\node[parallelepiped,draw=black!40,fill=columbiablue,
  minimum width=2cm,minimum height=0.01cm,parallelepiped offset x=9mm] (3) at (9.9,-1.5){};
\node[parallelepiped,draw=black!40,fill=brightube,
minimum width=4.93cm,minimum height=0.00051cm,scale=0.65,parallelepiped offset x=3mm,parallelepiped offset y=2mm] (7) at (10.28,-0.4){CNN};
\node[parallelepiped,draw=black!40,fill=columbiablue,
  minimum width=2cm,minimum height=0.01cm,parallelepiped offset x=9mm] (4) at (9.9,-3.7){};
\node [fill=white,scale=0.3,regular polygon,regular polygon sides=3, draw=black!70, shape border rotate=180] (16) at (9.88,-3.2) {};
\node [fill=hotpink,scale=0.3,regular polygon,regular polygon sides=3, draw=black!70, shape border rotate=180] (17) at (10.8,-3.4) {};
\node [fill=hotpink,scale=0.3,regular polygon,regular polygon sides=3, draw=black!70, shape border rotate=180] (18) at (11.3,-3.2) {};  
\node [fill=white,scale=0.26,star,star point ratio= 3,draw=black!70] (10) at (9.6,-1.2) {};
\node [fill=yellow,scale=0.26,star,star point ratio= 3,draw] (11) at (11.1,-1.05) {};
\node [fill=yellow,scale=0.26,star,star point ratio= 3,draw] (15) at (10.7,-1.09) {};
\node[parallelepiped,draw=black!40,fill=columbiablue,
  minimum width=2cm,minimum height=0.01cm,parallelepiped offset x=9mm] (5) at (13.3,-1.5){};
\node[parallelepiped,draw=black!40,fill=columbiablue,
  minimum width=2cm,minimum height=0.01cm,parallelepiped offset x=9mm] (6) at (13.3,-3.7){};
\node [fill=white,scale=0.3,regular polygon,regular polygon sides=3, draw=black!70, shape border rotate=180] (28) at (13.28,-3.2) {};
\node [fill=hotpink,scale=0.3,regular polygon,regular polygon sides=3, draw=black!70, shape border rotate=180] (29) at (14.2,-3.4) {};
\node [fill=hotpink,scale=0.3,regular polygon,regular polygon sides=3, draw=black!70, shape border rotate=180] (30) at (14.7,-3.2) {};  
\node [fill=white,scale=0.26,star,star point ratio= 3,draw=black!70] (19) at (13,-1.2) {};
\node [fill=yellow,scale=0.26,star,star point ratio= 3,draw] (20) at (14.4,-1.05) {};
\node [fill=yellow,scale=0.26,star,star point ratio= 3,draw] (21) at (14,-1.09) {};
\node [minimum width=2.5cm,minimum height=1cm,fill=green!20,draw=green!40,decorate,decoration={random steps, amplitude=1.6pt}] at (10.4,-2.35){};
\node [fill=white,scale=0.26,star,star point ratio= 3,draw=black!70] (22) at (9.6,-2.2) {};
\node [fill=yellow,scale=0.26,star,star point ratio= 3,draw] (23) at (11.1,-2.05) {};
\node [fill=yellow,scale=0.26,star,star point ratio= 3,draw] (24) at (10.7,-2.09) {};
\node [fill=white,scale=0.3,regular polygon,regular polygon sides=3, draw=black!70, shape border rotate=180] (25) at (9.88,-2.3) {};
\node [fill=hotpink,scale=0.3,regular polygon,regular polygon sides=3, draw=black!70, shape border rotate=180] (26) at (10.8,-2.5) {};
\node [fill=hotpink,scale=0.3,regular polygon,regular polygon sides=3, draw=black!70, shape border rotate=180] (27) at (11.3,-2.3) {};  
\draw[-Triangle,line width=.008cm,draw=black!80] (10.9,-1.16) -- (10.9,-1.9);
\draw[-Triangle,line width=.008cm,draw=black!80] (11,-3.35) -- (11.1,-2.4);
\draw[-Triangle,line width=.008cm,draw=black!80] (14.35,-3.3) -- (14.2,-1.2);
\draw[-Triangle,line width=.008cm,draw=black!80] (7.4,-1.2) -- (7.68,-3.2);
\draw (10.3,-0.54) edge[-Triangle,dashed,draw=black!70,out=-20,in=80] (14.2,-1.03);
\draw (10.3,-0.54) edge[-Triangle,dashed,draw=black!70,out=-10,in=80] (10.9,-1.02);
\draw (10.3,-0.54) edge[-Triangle,dashed,draw=black!70,out=200,in=80] (7.4,-1.05);
\node (rect) at (17,-4.0) [fill=black!10,text width=3.8cm,draw=black!20,minimum width=1cm,minimum height=0.8cm,scale=0.75] {Foggy effects;\\ White spots;\\ Dense inflammations;\\ Bright-pixels dominance};
\hspace*{4.9cm}\includegraphics[width=0.21\linewidth]{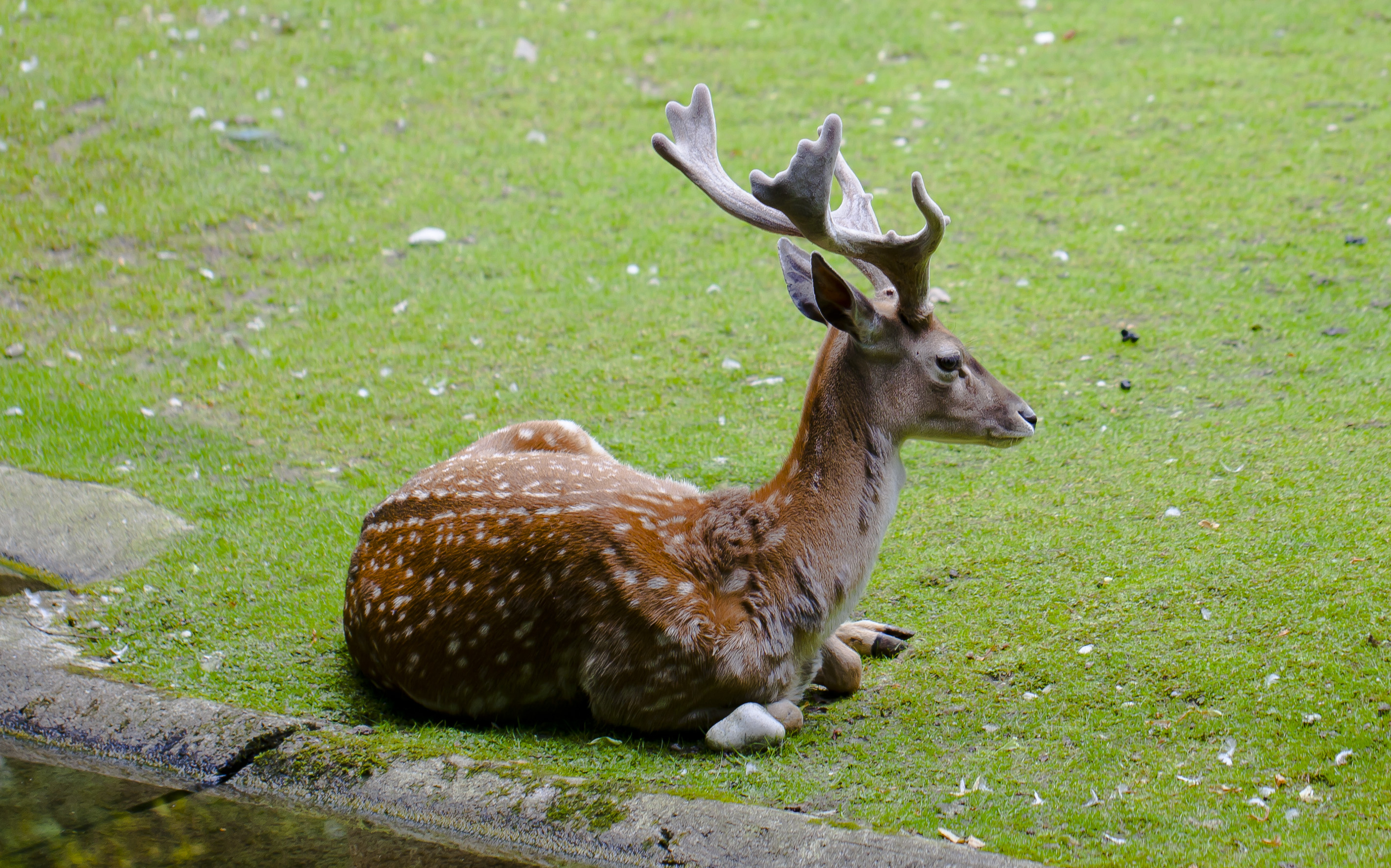};
\hspace*{0.3cm}\includegraphics[width=0.159\linewidth]{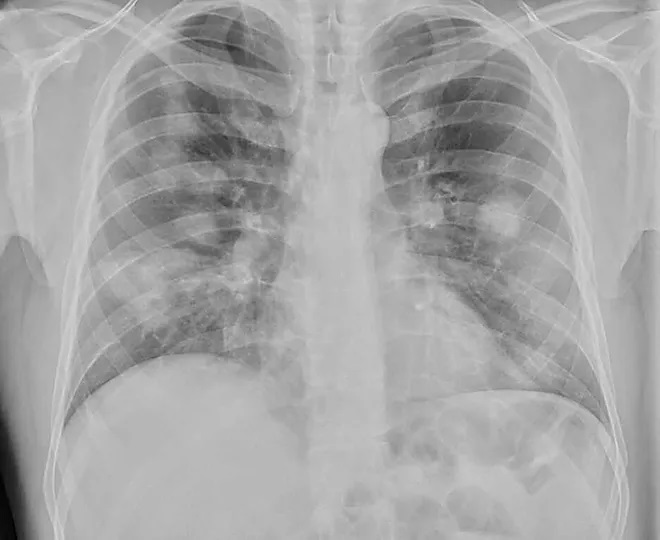};
\hspace*{0.3cm}\includegraphics[width=0.233\linewidth]{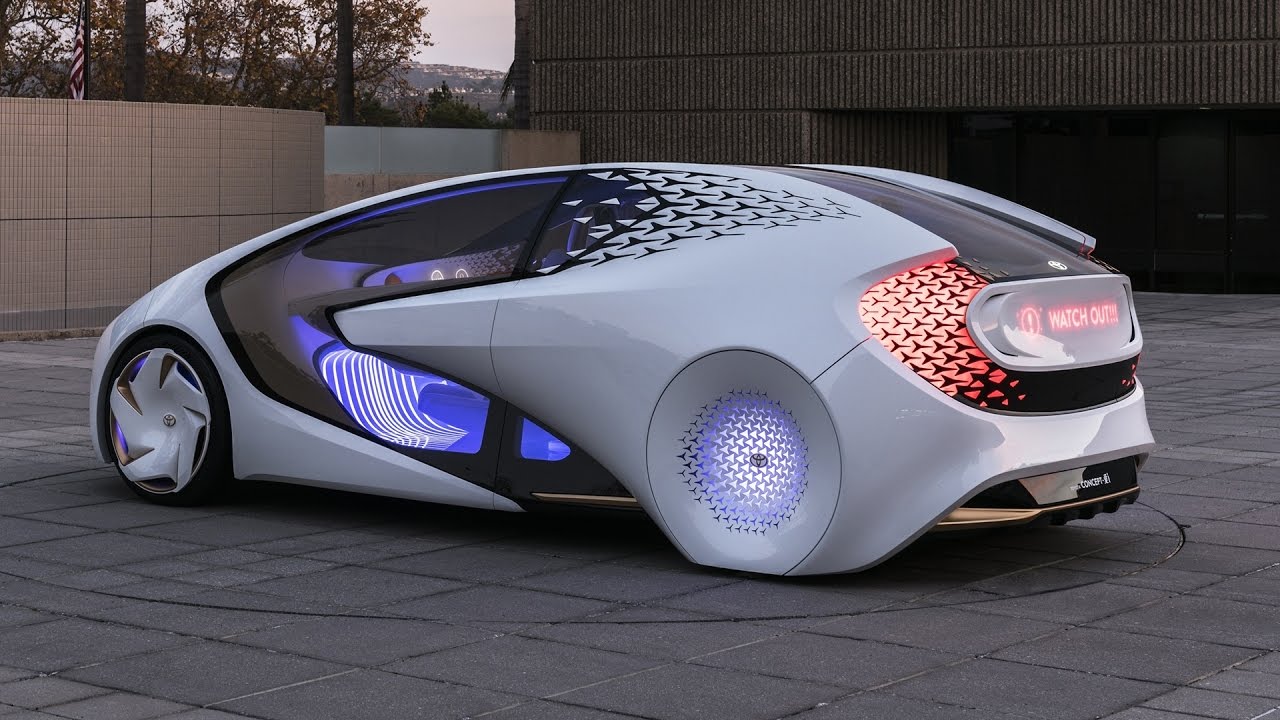};
\end{tikzpicture}%
\captionsetup{skip=0pt}
\vspace{3mm}
\caption{Overview of ZSL models. Typical approaches use one of the three embedding types or a combination of them. (a) Semantic embedding models that map visual features to the semantic space. (b) Models that map visual and semantic features to an intermediate latent space. (c) Visual embedding models that map semantic features to the visual space.} 
\label{fig:overview}
\end{figure*}

According to \cite{ye2017zero}, any approach that relies on label propagation will fall into the category of transductive learning. Feature generating network with labelled source data and unlabelled target data \cite{xian2019f} are also considered as transductive methods. The transductive setting is seen as one of the solutions to the domain shift problem, since the provided unseen labelled information during training reduces the discrepancy between the two domains.

There is a slight nuance between the transductive learning and semi-supervised learning; in the transductive setting, the unlabelled data solely belong to the unseen test classes, while in semi-supervised setting, unseen test classes might not be present in the unlabelled data. Furthermore, the difference between FSL and the transductive ZSL learning is the existence of a few labelled examples of the unseen classes alongside annotated seen class examples in the few-shot learning. While in the transductive ZSL setting, the examples for the unseen classes are all unlabelled.

ZSL models are developed based on two high-level major strategies to be taken into account: a) defining the \textit{``Embedding Space''} to combine visual and non-visual auxiliary data, and b) choosing an appropriate \textit{``Auxiliary Data Collection''} technique.\\

\vspace{-2mm}
\textit{a) Embedding Spaces}. Figure~\ref{fig:overview} demonstrates the overall structure of a ZSL system in terms of embedding spaces and auxiliary data types collection techniques. Such systems either map the visual data to the semantic space (Figure~\ref{fig:overview}.a) or embed both visual and semantic data to a common latent space (Figure~\ref{fig:overview}.b), or see the task as a missing data problem, and then map the semantic information to the visual space (Figure~\ref{fig:overview}.c). Two or all of these approaches can also be combined and embedded together to boost up the benefits of each individual categories.

From a different point of view, semantic spaces can also be sub-categorised into euclidean and non-euclidean spaces. The intrinsic relationship between data points is better preserved when the geometrical relation between them is considered. These spaces are commonly based on clusters or graph networks. Some researchers may prefer manifold learning for the ZSL challenge. e.g. in \cite{rohrbach2013transfer}, \cite{wang2016relational}, \cite{zhao2017zero}, \cite{xu2017transductive}, \cite{xu2017matrix},  \cite{li2017zeropath}, \cite{wang2018zero}, \cite{lee2018multi}, \cite{kampffmeyer2019rethinking}, \cite{zhu2019generalized}. The Euclidean spaces are more conventional and simpler as the data has a flat representation in such spaces. However, the loss of information is a common issue of these spaces, as well. Examples of methods using Euclidean spaces are \cite{lampert2013attribute}, \cite{frome2013devise}, \cite{romera2015embarrassingly}, \cite{xian2018feature}, \cite{mishra2018generative}, and \cite{schonfeld2019generalized}.\\

\vspace{-2mm}
\noindent \textit{b) Auxiliary Data Collection}. \noindent As mentioned before, Zero-shot learning is the challenge of learning novel classes without seeing their exemplars during the training. Instead, the freely available auxiliary information is used to compensate for the lack of visually labelled data. Such information can be categorised into two groups:

\textit{Human annotated attributes.} The supervised way of annotating each image with its related attributes is an arduous process and requires time and expertise, but since they are manual, they yield noiseless and important attributes needed for learning and inference. There are several datasets in which side information in the form of attributes can be attained for each image. i.e. aPY \cite{farhadi2009describing}, AWA1 \cite{lampert2013attribute}, AWA2 \cite{xian2017zero}, CUB \cite{wah2011caltech}, and SUN \cite{patterson2012sun}. Several ZSL methods leverage the attributes as the side information \cite{akata2015evaluation}, \cite{romera2015embarrassingly}, \cite{long2017zero}, or visual attributes \cite{lampert2009learning}, \cite{farhadi2009describing}.

\textit{Unsupervised auxiliary information.} There are several forms of auxiliary information that have minimum supervision and are widely used in the ZSL setting, such as human gazes \cite{karessli2017gaze}, WordNet which is a large-scale lexical database of 117,000 English words \cite{rohrbach2010helps}, \cite{rohrbach2011evaluating}, \cite{akata2013label}, \cite{akata2015evaluation}, \cite{xian2016latent}, \cite{lu2015unsupervised}, \cite{akata2016multi}, \cite{qiao2016less}, \cite{marino2016more}, \cite{wang2018zero}, \cite{lee2018multi}, or Textual descriptions such as Web search \cite{rohrbach2011evaluating}, Wikipedia articles \cite{frome2013devise}, \cite{norouzi2013zero}, \cite{elhoseiny2013write}, \cite{akata2015evaluation}, \cite{lei2015predicting}, \cite{akata2016multi}, \cite{qiao2016less}, \cite{elhoseiny2017link}, \cite{niu2018webly}, \cite{zhu2018generative}, and sentence descriptions \cite{reed2016learning}. Textual side information needs to be transformed into class embeddings in order to be used at the training stage and testing stages. Word embedding and language embedding are the two representation techniques used for textual side information. As we gradually proceed, later we review on different embedding classes as well.

\section{ZSL Data Embedding Techniques\label{appro}}
\noindent In this section, we first provide the task definition of ZSL and GZSL. Then we review the four recent approaches on the problem. 

In the standard inductive setting as mentioned earlier in Section \ref{ZSL_TT}, the training set is
\begin{equation}
D_{t}=\{(x,y,c(y))|x\in X^S,y\in Y^{S}, c(y) \in C^S\}	
\end{equation}

\noindent and the objective function to be minimised is as follows: 
\begin{align}
\mathpzc {L} = \frac{1}{N}\displaystyle \sum^N_{n=1}L(y_n,f(x_n;W))+\Omega(W)
\end{align}
where, $f(x,y;W) = \underset {y \in Y} \argmax F(x,y)$ is the mapping function.\\Through the training phase, the classifier $f:X\rightarrow Y^{U}$ is learned for ZSL to predict only the novel classes at the test time, or $f:X\rightarrow Y^{U}\cup Y^{S}$ for the GZSL challenge to estimate both novel classes and the previously learned seen classes. For instance, the classifier $f$ can be a COVID-19 diagnoser.\\

We categorise the embedding methodologies into four categories based on the space they learn/infer target classes (like COVID-19 detection in Figure \ref{fig:overview}):

\begin{enumerate}
	\item Semantic Embedding: A semantic space with textual nature in which features are in the form of class embeddings.
	\item Intermediate-Space Embedding: A space where both class embeddings and visual feature embedding present in conjunction.
	\item Visual Embedding: A space where training and inferring is done with visual feature representations similar to the traditional recognition problems.
	\item Hybrid Embedding Models: A combination of spaces are used in some models to bring together the advantages the different spaces have.
\end{enumerate}
The majority of methods focus on the general tasks; however, they are scalable to disease classification. 

\subsection{Semantic Embedding}
Semantic embedding itself can be sub categorised into two tasks of \textit{Attribute Classification} and \textit{Label Embedding} which will be discussed here:
 
\subsubsection{Attribute Classifiers}
Primitive approaches of Zero-Shot learning leverage manually annotated attributes in a two-stage learning schema. Attributes in an image are predicted in the first stage and labels of unseen classes would be chosen using similarity measures in the second stage. \cite{lampert2009learning} use a probabilistic classifier to learn the attributes and then estimates posteriors for test classes. \cite{rohrbach2010helps} propose a method to avoid manual supervision with mining the attributes in an unsupervised manner. \cite{rohrbach2011evaluating} adopt DAP together with a hierarchy-based knowledge transfer for large-scale settings. \cite{kankuekul2012online}'s method is based on IAP, and uses Self-Organising and Incremental Neural Networks (SOINN) to learn and update attributes online. Later in IAP-SS by \cite{kankuekul2012online}, an online incremental learning approach is used for faster learning of the new attributes.  The Direct Attribute Prediction (DAP) \cite{lampert2013attribute} first learns the posteriors of the attributes, then estimates the posteriors of seen classes. On the other hand, Indirect Attribute Prediction (IAP) \cite{lampert2013attribute} first learns the posteriors for seen classes then uses them to compute the posteriors for the attributes. \cite{wang2013unified} use a unified probabilistic model based on the Bayesian Network (BN) \cite{murphy2012machine} that discovers and captures both object-dependent and object-independent relationships to overcome the problem of relating the attributes. ConSE \cite{norouzi2013zero} learns the probability of the training samples. It then predicts an unseen class by the convex combination of the class label embedding vectors.
\cite{jayaraman2014zero} use a random forest approach for learning more discriminative attributes. Hierarchy and Exclusion (HEX) \cite{deng2014large} considers relations between objects and attributes and maps the visual features~\cite{mtei2019cnn}, ~\cite{Rezaei2014b} of the images to a set of scores to estimate labels for unseen categories. \cite{al2016recovering} take on an unsupervised approach where they capture the relations between the classes and attributes with a three-dimensional tensor while using a DAP-based scoring function to infer the labels. LAGO by \cite{atzmon2018probabilistic} also follow the DAP model. It learns soft and-or logical relations between attributes. Using soft-OR, the attributes are divided into groups, and the label class from unseen samples is predicted via a soft-AND within these groups. If each attribute comes from a singleton group, the all-AND will be used.

\subsubsection{Label Embedding}
\noindent Instead of using an intermediate step, more recent approaches learn to map images to the structured euclidean semantic space automatically which would be the implicit way of representing knowledge. The compatibility function for linear mapping is:
\begin{align}
F(x,y;w)=\theta (x)^Twc(y)
\end{align}

\noindent where $\theta (x)^T$ is the image embedding for training classes and $w$ is the parameters in vector form to be learned. In the case of bilinear projection where it is more common, $w$ takes the form of matrix:
\begin{align}
F(x,y;W)=\theta (x)^TWc(y)
\end{align}
SOC \cite{palatucci2009zero} first maps the image features to the semantic embedding space, it then estimates the correct class using nearest neighbour. DeViSE by \cite{frome2013devise} uses a linear corresponding function with a combination of dot-product similarity and hinge rank loss used in \cite{weston2010large}. ALE \cite{akata2015label} optimises the ranking loss in \cite{usunier2009ranking} alongside the bi-linear mapping compatibility function. SJE \cite{akata2015evaluation} learns a bi-linear compatibility function using the structural SVM objective function \cite{tsochantaridis2005large}. ESZSL \cite{romera2015embarrassingly} introduces a better regulariser and optimises a close form solution objective function in a linear manner. ZSLNS \cite{qiao2016less} proposes a $l_{\mathrm{1,2}}$-norm based loss function. \cite{bucher2016improving} take on a metric learning approach and linearly embed the visual features to the attribute space. LAGO \cite{atzmon2018probabilistic} is a probabilistic model that depicts soft and-or relations between groups of attributes. In a case where all attributes form all-OR group, It becomes similar to ESZSL \cite{romera2015embarrassingly} and learns a bilinear compatibility function. AREN \cite{xie2019attentive} uses attentive region embedding while learning the bilinear mapping to the semantic space in order to enhance the semantic transfer. ZSLPP \cite{elhoseiny2017link} combines two networks VPDE-net for detecting bird parts from images and PZSC-net that trains a part-based Zero-Shot classifier from the noisy text of the Wikipedia. DSRL \cite{ye2017zero} uses non-negative sparse matrix factorisation to align vector representations with the attribute-based label representation vectors so that more relevant visual features are passed to the semantic space.

Some approaches to ZSL use non-linear compatibility functions. CMT \cite{socher2013zero} use a two-layer neural network, similar to common MLP networks by \cite{rezaei2018a} alongside the compatibility function. In UDA \cite{kodirov2015unsupervised} a non-linear projection from feature space to semantic space (word vector and attribute) is proposed in an unsupervised domain adaptation problem based on regularised sparse coding. \cite{lei2015predicting} use a deep neural network~\cite{mtei2019cnn} regression which generates pseudo attributes for each visual category via Wikipedia. LATEM \cite{xian2016latent} constructs a piece-wise non-linear compatibility function alongside a ranking loss. \cite{changpinyo2017predicting} regularise the model using structural relations of the cluster by which cluster centres characterise visual features. QFSL by \cite{song2018transductive} solves the problem in a transductive setting, and projects both sources and target images into several specified points to fight bias problem.

GFZSL \cite{verma2017simple} uses both linear and non-linear regression models and generates a probability distribution for each class. For transductive setting, it uses Expectation-Maximisation (EM) to estimate a Gaussian Mixture Model (GMM) of unlabelled data in an iterative manner.

Leveraging the non-euclidean spaces to capture the manifold structure of the data is another approach to the problem. Together with the knowledge graphs, the explicit relations between the labels will be demonstrated. In this setting, the side information mainly comes from a hierarchy ontology like WordNet. The mapping function will have the following form:
\begin{align}
F(x,y;W)=\theta (X,A)^TWc(y)
\end{align}

\noindent where $X$ is the $n \times k$ feature matrix and $A$ is the adjacency matrix of the graph.\\ Propagated Semantic Transfer (PST) \cite{rohrbach2013transfer} first uses DAP model to transfer knowledge to novel categories, following the graph-based learning schema, it improves local neighbourhood in them. DMaP \cite{li2017zeropath} jointly optimises the projecting of the visual features and the semantic space to improve the transferability of the visual features to the semantic space manifold. MFMR \cite{xu2017matrix} decomposes the visual feature matrix into three matrices to further facilitate the mapping of visual features to the semantic spaces. To improve the representation of the geometrical manifold structure of the visual and semantic features, manifold regularisation is used. In \cite{lee2018multi} a Graph Search Neural Network (GSNN) \cite{marino2016more} is used in the semantic space based on the WordNet knowledge graph to predict multiple labels per image using the relations between them. \cite{wang2018zero} distils both auxiliary information in forms of word embedding and knowledge graph to learn novel categories. DGP \cite{kampffmeyer2019rethinking} proposes dense graph propagation to propagate knowledge directly through dense connections. In \cite{zhu2019generalized} a graphical model with a low dimensional visually semantic space is utilised which has a chain-like structure to close the gap between the high-dimensional features and the semantic domain.

\subsection{Intermediate-Space Embedding}

One of the methods of embedding is to measure the similarity between the visual and semantic features in a joint space.

\subsubsection{Fusion-based Models}
Considering unseen classes as a fusion of previously learned seen concepts is called hybrid learning. Standard scoring function for hybrid models is defined as:
\begin{align}
f(x,y;W) &= \underset {s \in S} \sum (W,\theta ^{S}(x))c(y)                              
\end{align} 
SSE \cite{zhang2015zero} considers the histogram similarity between the seen class auxiliary information and seen visual data.
In SYNC \cite{changpinyo2016synthesized} uses two spaces of semantic and model space, and the alignment is conducted with phantom classes. With the sparse linear combination of the classifiers for the phantom classes, the final classifier is learned. TVSE \cite{xu2017transductive} learns a latent space using collective matrix factorisation with graph regularisation to incorporate the manifold structure between source and target instances, moreover, it represents each sample as a mixture of seen class scores. LDF \cite{li2018discriminative} combines the prototypes of seen classes and jointly learns embeddings for both user-defined attributes and latent attributes.
\subsubsection{Joint Representation Space Models}
Inferring unseen labels via measuring similarity between cross-modal data in a shared latent space is another workaround to the ZSL challenge. The first term in the objective function for standard cross-modal alignment approaches is:
\begin{align}
\mathpzc {L} = \underset {c(y)^{S}} \min \| x^{S} - c(y)^{S} y^{S} \| ^2 _F
\end{align}
with $Y$ being a One-hot vector of corresponding class labels and $\| . \| ^2_F$ is the Frobenius norm. Approaches to joint space learning are grouped into two categories, Parametric which follow a slow learning via optimising a problem and Non-parametric that leverage data points extracted from neural networks in a shared space. In parametric methods including \cite{fu2015transductive} a multi-view alignment space is proposed for embedding low-level visual features. The learning procedure is based on the multi-view Canonical Correlation Analysis (CCA) \cite{gong2014multi}. \cite{lu2015unsupervised} applies PCA and ICA embeddings to reveal the visual similarity across the classes and obtains the semantic similarity with the WordNet graph, followed by embedding the two outputs into a common space.
MCZSL \cite{akata2016multi} uses visual part and multi-cue language embedding in a joint space.
In \cite{mukherjee2016gaussian} both images and words are represented by Gaussian distribution embeddings. JLSE \cite{zhang2016zero} decides on a dictionary learning approach to learn the parameters of source and target domains across two separate latent spaces where the similarity is computed by the likelihood of similarity independent to the class label. CDL \cite{jiang2018learning} uses a coupled dictionary to align the structure of visual-semantic space using discriminative information of the visual space. In \cite{kolouri2018joint} and \cite{rostami2019zero} a coupled sparse dictionary is leveraged to relate visual and attribute features together. It uses entropy regularisation to alleviate the domain shift problem.

There are several non-parametric methods. ReViSE \cite{tsai2017learning} that combines auto-encoders with Maximum Mean Discrepancy (MMD) loss \cite{gretton2007kernel} in order to align the visual and textual features. DMAE \cite{mukherjee2017deep} introduces a latent alignment matrix with representations from auto-encoders optimised by kernel target alignment (KTA) \cite{cristianini2002kernel} and squared-loss mutual information (SMI) \cite{yamada2015cross}. DCN \cite{liu2018generalized} proposes a novel Deep Calibration Network in which an entropy minimisation principle is used to calibrate the uncertainty of unseen classes as well as seen classes.

To narrow the semantic gap, BiDiLEL \cite{wang2017zero} introduces a sequential bidirectional learning strategy and creates a latent space using the visual data, then the semantic representations of unseen classes are embedded in the previously created latent space. This method comprises both parametric and non-parametric models.

\subsection{Visual Embedding}
\noindent Visual embedding is the other type of ZSL methods that performs classification in the original feature space and is orthogonal to semantic space projection. This is done by learning a linear or non-linear projection function. For linear corresponding functions, WAC-Linear \cite{elhoseiny2013write} uses textual description for seen and unseen categories and projects them to the visual feature space with a linear classifier. \cite{zhao2017zero} follows a transductive setting in which it refines unseen data distributions using unseen image data. To approximate the manifold structure of data, they used a global linear mapping for synthesising virtual cluster centres. \cite{guo2017zero} assigns pseudo labels to samples using reliability (with robust SVM) and diversity (via diversity regularisation).
For learning  Ia Non-linear corresponding function, In WAC-Kernel \cite{elhoseiny2016write} in order to leverage any kind of side information, a kernel method is proposed to predict a kernel-based on the representer theorem \cite{scholkopf2001generalized}. DEM \cite{zhang2017learning} uses the least square embedding loss to minimise the discrepancy between the visual features and their class representation embedding vector in the visual feature space. OSVE \cite{long2017towards} reversely maps from attribute space to visual space then trains the classifier using SVM~\cite{arvanaghi2017a}. In~\cite{ji2018stacked} the authors introduce a stacked attention network that corporates both global and local visual features weighted by relevance along with the semantic features. In \cite{wan2019transductive} visual constraint is used in class centres in the visual space to avoid the domain shift problem.
\subsubsection{Visual Data Augmentation}
There are a variety of generative networks that augment unseen data, taking GAN \cite{goodfellow2014generative} as an example, the first term in objective function would be:
\begin{align}
\mathpzc {L} = \max \mathbb{E}[\log D(x,c(y))]+\min \mathbb{E}[\log (1-D(\tilde {x},c(y))]
\end{align}
$\tilde {x}=G(z,c(y))$ is the synthesised data of the generator and $z \in R^{d_z}$ is random Gaussian noise.
The role of the discriminator $D$ and generator $G$ contradicts in loss function as the first one attempts to maximise the loss while the latter tries to minimise it.\\
Another widely used generative neural network is the Variational AutoEncoder (VAE) \cite{kingma2013auto}:
\begin{align}
\mathpzc {L} = \mathbb{E}_{q_\phi (z|x)}[\log p_{\theta}(x|z)]-D_{KL}(q_{\phi}(z|x)||p_{\theta}(z))
\end{align}
The first term is the reconstruction loss, and the latter is the Kullback-Leibler divergence that works as a regulariser.

RKT \cite{wang2016relational} leverages relational knowledge of the manifold structure in the semantic space, and generates
virtually labelled data for unseen classes from Gaussian distributions generated by sparse coding. Then it projects them alongside the seen data to the semantic space via linear mapping. 
GLaP \cite{li2017zero} generates virtual instances of an unseen class with the assumption that each representation obeys a prior distribution where one can draw samples from. To ease the embedding to the semantic space, GANZrl \cite{tong2018adversarial} proposes to increase the visual diversity by generating samples with specified semantics using GAN models. 
SE-GZSL \cite{kumar2018generalized} uses a feedback-driven mechanism for its discriminator that learns to map the produced images to the corresponding class attribute vectors. To enforce the similarity of the distribution of the sample and generated sample, a loss component was added to the VAE objective \cite{kingma2013auto} function.

Synthesised images often suffer from looking unrealistic since they lack intricate details. A way around this issue is to generate features instead.
\cite{bucher2017generating} uses a GMMN model \cite{li2015generative} to generate visual features for unseen classes. In \cite{felix2018multi} a multi-modal cycle consistency loss is used in training the generator for better reconstruction of the original semantic features. CVAE-ZSL \cite{mishra2018generative} takes attributes and generates features for the unseen categories via a Conditional Variations AutoEncoder (CVAE) \cite{sohn2015learning}. $L_2$ norm is used as the reconstruction loss. 
GAZSL \cite{zhu2018generative} utilises noisy textual descriptions from Wikipedia to generate visual features. A visual pivot regulariser is introduced to help generate features with better qualities.
f-CLSWGAN \cite{xian2018feature} combines three conditional GAN variants for a better data generation.
f-VAEGAN-D2 \cite{xian2019f} combines the architectures of conditional VAE \cite{sohn2015learning}, GAN \cite{goodfellow2014generative} and a non-conditional discriminator for the transductive setting. LisGAN \cite{li2019leveraging} generates unseen features from random noises using conditional Wasserstein GANs \cite{arjovsky2017wasserstein}. For regularisation, they introduced semantically meaningful soul samples for each class and forced the generated features to be close to at least one of the soul samples. Gradient Matching Network (GMN) \cite{sariyildiz2019gradient} trains an improved version of the conditional WGAN \cite{gulrajani2017improved} to produce image features for the novel classes. It also introduces Gradient Matching (GM) loss to improve the quality of the synthesised features. In order to synthesise unseen features, SPF-GZSL \cite{li2019generalized} selects similar instances and combines them to form pseudo features using a centre loss function \cite{wen2016discriminative}. In Don’t Even Look Once (DELO) by \cite{zhu2019dont} a detection algorithm is conducted to synthesise unseen visual features to gain high confidence predictions for unseen concepts while maintaining low confidence for backgrounds with vanilla detectors.

Instead of augmenting data using synthesising methods, data can be acquired by gathering web images. \cite{niu2018webly} jointly use web data which are considered weakly-supervised categories alongside the fully-supervised auxiliary labelled categories. It then learns a dictionary for the two categories.

\begin{table*} [hbt!]
		\footnotesize
		\renewcommand\arraystretch{3}
    \centering
    \caption{Common ZSL and GZSL methods categorised based on their embedding space model, with further divisions in a top-down manner.\\}
		\hspace{1mm}
    \begin{tabular}{ c | c | c | l }                   
\hline
\textbf{Models}
& \textbf{Categories}     
    &   \textbf{Main Features}         
        &    \textbf{Description}      
\\
 \hline
	\multirow{3}{17mm}{\centering \rule{0pt}{0ex}Semantic Embedding}
    &   \multirow{1}{20mm}{\centering \rule{0pt}{0ex}Two-Step Learning}
        & \multirow{1}{36mm}{\centering \rule{0pt}{0ex}Attributes classifiers }
            & \multirow{1}{88mm}{\rule{0pt}{3ex}DAP-Based \cite{lampert2009learning}, \cite{rohrbach2010helps}, \cite{rohrbach2011evaluating}, \cite{lampert2013attribute}, \cite{al2016recovering}, \cite{atzmon2018probabilistic} IAP-Based \cite{lampert2009learning}, \cite{kankuekul2012online}, \cite{lampert2013attribute}, \cite{norouzi2013zero} Bayesian network (BN) \cite{wang2013unified}, Random Forest Model \cite{jayaraman2014zero}, HEX Graph \cite{deng2014large}}
\\ 
	\cline{2-4}
    &    \multirow{2}{20mm}{\rule{0pt}{0ex}Direct Learning}
        &   \multirow{1}{36mm}{\centering \centering \rule{0pt}{0ex}Implicit knowledge representation}
            & \multirow{1}{88mm}{\rule{0pt}{0ex}Linear \cite{palatucci2009zero}, \cite{frome2013devise}, \cite{akata2015label}, \cite{akata2015evaluation}, \cite{romera2015embarrassingly}, \cite{qiao2016less}, \cite{bucher2016improving}, \cite{atzmon2018probabilistic}, \cite{xie2019attentive}, \cite{elhoseiny2017link}, \cite{ye2017zero}, \cite{verma2017simple} or Non-Linear \cite{kodirov2015unsupervised}, \cite{lei2015predicting}, \cite{xian2016latent}, \cite{changpinyo2017predicting}, \cite{song2018transductive}, \cite{verma2017simple}Compatibility Functions}
\\ 
   \cline{3-4}
            & 
            &\multirow{1}{36mm}{\centering \centering \rule{0pt}{0ex}Explicit knowledge representation}
            & \multirow{1}{88mm}{\rule{0pt}{0ex}Graph Conv. Networks (GCN)~\cite{wang2018zero}, Knowledge Graphs\cite{lee2018multi}, \cite{rohrbach2013transfer}, \cite{li2017zeropath}, \cite{kampffmeyer2019rethinking}, 3-Node Chains~\cite{zhu2019generalized}, Matrix Tri-Factorisation with Manifold Regularisation \cite{xu2017matrix}}
\\
  \hline
\multirow{2}{17mm}{ \centering \rule{0pt}{0ex}Cross-Modal\, Latent Embedding } 
    &   \multirow{1}{20mm}{\centering Fusion-based Models}
        & \multirow{1}{36mm}{\centering Fusion of seen class data}
            & \multirow{1} {88mm}{\rule{0pt}{0ex}Combination of seen classes properties \cite{zhang2015zero}, \cite{changpinyo2016synthesized}, \cite{li2018discriminative}, Combination of seen class scores \cite{xu2017transductive}} 
\\ 
  \cline{2-4}
	  &   \multirow{1}{20mm}{\centering \rule{0pt}{0ex}Common Representation Space Models}
       	&   \multirow{1}{36mm}{\centering \rule{0pt}{0ex}Mapping of the visual and semantic spaces in a joint intermediate space}
            & \multirow{1}{88mm}{\rule{0pt}{0ex}Parametric \cite{fu2015transductive}, \cite{lu2015unsupervised}, \cite{akata2016multi}, \cite{mukherjee2016gaussian}, \cite{zhang2016zero}, \cite{jiang2018learning}, \cite{kolouri2018joint}, \cite{rostami2019zero}, Non-parametric \cite{tsai2017learning}, \cite{mukherjee2017deep}, \cite{liu2018generalized}, or Both  \cite{wang2017zero} }  
\\
  \hline 
\multirow{4}{17mm}{\centering \rule{0pt}{0ex}Visual Embedding}
    &   \multirow{1}{20mm}{\centering \centering \rule{0pt}{0ex} Visual Space Embedding} 
        &  \multirow{1}{36mm}{\centering \centering \rule{0pt}{0ex} Learning of the semantic to visual projection}
            & \multirow{1}{88mm}{\rule{0pt}{0ex}Linear \cite{elhoseiny2013write}, \cite{zhao2017zero}, \cite{guo2017zero} or Non-linear \cite{elhoseiny2016write}, \cite{zhang2017learning}, \cite{long2017towards}, \cite{ji2018stacked}, \cite{wan2019transductive} Projection functions}
\\
  \cline{2-4}
    &   \multirow{2}{20mm}{\centering \rule{0pt}{5ex} Data Augmentation}
        &   \multirow{1}{36mm}{\centering Image generation}
        & \multirow{1}{88mm}{Gaussian distribution \cite{wang2016relational}, \cite{li2017zero}, GAN \cite{tong2018adversarial}, VAE \cite{kumar2018generalized}} 
	\\
   \cline{3-4}
            & 
            &   \multirow{1}{36mm}{\rule{0pt}{0ex} Visual feature generation}             
            &  \multirow{1}{88mm}{\rule{0pt}{0ex}GAN \cite{felix2018multi}, \cite{zhu2018generative},  \cite{li2019leveraging}, WGAN \cite{xian2018feature}, \cite{sariyildiz2019gradient}, CVAE \cite{mishra2018generative}, \cite{zhu2019dont}, VAE+GAN \cite{xian2019f}, GMMN \cite{bucher2017generating}, Similar feature combination \cite{li2019generalized}} 
\\
  \cline{2-4}
    &   \multirow{1}{20mm}{\centering Leveraging Web Data}
        &   \multirow{1}{36mm}{\centering Web images crawling}
            &   \multirow{1}{88mm}{Dictionary learning \cite{niu2018webly}}  
\\               
  \hline
\multirow{3}{17mm}{\centering \rule{0pt}{0ex}Hybrid}  
    &    \multirow{1}{20mm}{\centering Visual+Semantic Embedding}
        & \multirow{1}{36mm}{\centering Reconstruction of the semantic features}
            &   \multirow{1}{88mm}{\rule{0pt}{0ex}AutoEncoder \cite{kodirov2017semantic}, Adversarial AutoEncoder \cite{chen2018zero}, GAN with two reconstructing regressors \cite{shibing2019bi}, GAN an inverse GAN \cite{chen2019canzsl}} 
\\
  \cline{2-4}
    &   \multirow{1}{20mm}{\centering Visual+Cross Modal Embedding}
        &   \multirow{1}{36mm}{\centering Feature generation with aligned semantic features}
            &   \multirow{1}{88mm}{Semantic to visual mapping \cite{long2017zero}, VAE \cite{schonfeld2019generalized}} \\
  \cline{2-4}
    &  \multirow{1}{20mm}{\centering All}
        &   \multirow{1}{36mm}{\centering Utilisation of generator and discriminator together with the regressor}
            &   \multirow{1}{88mm}{GAN + Dual Learning \cite{huang2019generative}} 
\\               
  \hline
\end{tabular}
\label{tab:summary}
\end{table*}

\subsection{Hybrid Embedding Models} 
\noindent Several works make use of both visual and semantic projections to reconstruct better semantics to confront domain shift issue by alleviating the contradiction between the two domains. Semantic AutoEncoder (SAE) \cite{kodirov2017semantic} adds a visual feature reconstruction constraint. It combines linear visual-to-semantic (encoder) and linear semantic-to-visual (decoder). SP-AEN \cite{chen2018zero} is a supervised Adversarial AutoEncoder \cite{makhzani2015adversarial} which improves preserving the semantics by reconstructing the images from the raw 256 x 256 x 3 RGB colour space. BSR \cite{shibing2019bi} uses two different semantic reconstructing regressors to reconstruct the generated samples into semantic descriptions. CANZSL \cite{chen2019canzsl} combines feature-synthesis with semantic embedding by using a GAN for generating visual features and an inverse GAN to project them into semantic space. In this way, the produced features are consistent with their corresponding semantics.

Some of the synthesising approaches utilise a common latent space to align the generated features space with the semantic space to facilitate capturing the relations between the two spaces. \cite{long2017zero} introduce a latent-structure-preserving space where synthesised features from given attributes would suffer less from bias and variance decay with the help of Diffusion Regularisation. CADA-VAE \cite{schonfeld2019generalized} generates a visual feature latent space where both of visual and semantic features are embedded in this space by a VAE \cite{kingma2013auto}. It uses Distribution Alignment (DA) loss and Cross-Alignment (CA) loss to align the cross-modal latent distributions.

GDAN \cite{huang2019generative} combines all three approaches and designs a dual adversarial loss. 
In this way, regressor and discriminator learn from each other.

A summary of the different approaches is reported in Table~\ref{tab:summary}. The number of methods are growing with time and we can interpret that some areas like direct learning, common space learning and visual data synthesising are more popular in solving the task, while models combining different approach are fairly newer techniques thus have fewer works that are reported here.

\section{Evaluation Protocols\label{eval}}

In this section, we review some of the standard evaluation techniques to analyse the performance of the ZSL techniques based on the common benchmark datasets in the field, also in terms of dataset splits, class embeddings, image embeddings, and various evaluation metrics. First, the benchmark datasets.\\

\subsection{Benchmark Datasets}
\noindent There are several well-known benchmark datasets for Zero-shot learning that are frequently used. \\
\textbf{North America Birds (NAB)} \cite{van2015building} is a fine-grained dataset of birds consisting of 1,011 classes and 48,562 images. Images are categorised based on their visual attributes. A new version of this dataset is proposed by \cite{elhoseiny2017link} in which the identical leaf nodes are merged to their parent nodes where their only differences were genders and resulted in final 404 classes.\\

\begin{table*}[t]
\centering
\caption{Statics of the attribute datasets accounting for the number of attributes, classes plus their splits and their total number of images.\\}
\hspace{3mm}

    \begin{tabular}{l|ccccc} 
      \toprule 
       \multicolumn{1}{c|}{Attribute Datasets} & \#attributes & $y$ & $y^{U}$ & $y^{S}$ & \#images\\
      \midrule 
      SUN \cite{patterson2012sun} & 102 & 717 & 580+65 & 72 & 14,340\\    
      CUB \cite{wah2011caltech} & 312 & 200 & 100+50 & 50 & 11,788\\ 
      AWA1 \cite{lampert2013attribute} & 85 & 50 & 27+13 & 10 & 30,475\\ 
      AWA2 \cite{xian2017zero} & 85 & 50 & 27+13 & 10 & 37,322\\ 
      aPY \cite{farhadi2009describing} & 64 & 32 & 15+5 & 12 & 15,339\\          
      \bottomrule 
\end{tabular}
\label{tab:datasets}
\end{table*}

\textbf{Attribute datasets.} SUN Attribute \cite{patterson2012sun} is a medium-scale and fine-grained attribute dataset consisting of 102 attributes, 717 categories and a total of 14,340 images of different scenes.
CUB-200-2011 Birds (CUB) \cite{wah2011caltech} is a 200 category fine-grained attribute dataset with 11,788 images of bird species that includes 312 attributes.
Animals with Attributes (AWA1) \cite{lampert2013attribute} is another attribute dataset of 30,475 images with 50 categories 
and 85 attributes, the image features in this dataset are licensed and not available publicly. Later, Animals with Attributes 2 (AWA2) was presented by \cite{xian2017zero} which is a free version of AWA1 with more images than the previous one (37,322 images), with the same number of classes and attributes, but different images. 
aPascal and Yahoo (aPY) \cite{farhadi2009describing} is a dataset with a combination of 32 classes, including 20 pascal and 12 yahoo attribute classes with 15,339 images and 64 attributes in total.\\

A summary of the statics for the attribute datasets are gathered in Table~\ref{tab:datasets}.\\

\textbf{ImageNet.} ImageNet \cite{deng2009imagenet} is a large-scale dataset that contains 14 million images, shared between 21k categories with each image having one label that makes it a popular benchmark to evaluate models in real-world scenarios. Its organisation is based on WordNet hierarchy\cite{miller1995wordnet}. ImageNet is imbalanced between classes as the number of samples in each class vary greatly and is partially fine-grained. A more balanced version has 1k classes with 1000 images in each category.

There are several approaches in FSL setting for COVID-19 diagnosis, however ZSL is still new in the field of disease recognition, we introduce a dataset suited for the task of ZSL/GZSL that contains the required image and textual descriptions in one place.\\

\textbf{COVID-ChestXRay.} COVID-ChestXRay \cite{cohen2020covid} is a small and public dataset of CXR and CT scans suitable for ZSL and Few-shot learning experiences. At the time of this research, it had 444 unique clinical notes for a total of 16 categories, from no finding (normal cases) to other pneumonic cases like COVID-19, MERS, and SARS.

\subsection{Dataset Splits}
\noindent Here we discuss the original splits of the datasets as well as the other splits proposed for the Zero-shot problem.\\

\textbf{Standard Splits (SS).} In ZSL problems, unseen classes should be disjoint to seen classes and test time samples limited to unseen classes, thus the original splits aim to follow this setting. SUN \cite{patterson2012sun} proposed to use 645 classes for training among which 580 of the classes are used for training, 65 classes are for validation and the remaining 72 classes will be used for testing. For CUB, \cite{akata2015label} introduces the split of 150 training classes (including 50 validation classes) and 50 test classes. As for AWA1, \cite{lampert2013attribute} introduced the standard split of 40 classes for training (13 validation classes) and 10 classes for testing. The same splits are used for AWA2. In aPY, 20 classes of Pascal are used for training (15 classes for training and 5 for validation), while the 12 classes of Yahoo are used for testing.\\

\textbf{Proposed Splits (PS).} The standard split images from SUN, CUB, AWA1 and aPY overlap with some images of pre-trained ResNet-101 ImageNet model. To solve the problem, proposed splits (PS) is introduced by \cite{xian2018zero} where no test images are contained in the ImageNet 1K dataset.\\

\textbf{ImageNet.} \cite{xian2018zero} proposes 9 ZSL splits for the ImageNet dataset; two of which evaluate the semantic hierarchy in distance-wise scales of 2-hops (1509 classes) and 3-hops (7678 classes) from the 1k training classes. The remaining six splits consider the imbalanced size of classes with increasing granularity splits starting from 500, 1K and 5K least-populated classes to 500, 1K and 5K most-populated classes, or All which denotes a subset of 20k other classes for testing.\\

\textbf{Seen-Unseen relatedness.} To measure the relatedness of seen samples to unseen classes, \cite{elhoseiny2017link} introduces two splits Super-Category-Shared (SCS) and Super-Category-Exclusive (SCE). SCS is the easy split since it considers the relatedness to the parent category while SCE is harder and measures the closeness of an unseen sample to that particular child node. 

\subsection{Class Embeddings}
\noindent There exist several class embeddings, each suitable for a specific scenario. Class embeddings are in forms of vectors of real numbers which can further be used to make predictions based on the similarity between them and can be obtained through three categories: attributes, word embeddings, and hierarchical ontology. The last two are done in an unsupervised manner thus do not require human labour.

\subsubsection{Supervised Attribute-Embeddings}
Human annotated attributes are done under the supervision of experts with a great amount of effort. Binary, relative and real-valued attributes are three types of attributes embeddings. Binary attributes depict the presence of an attribute in an image thus value is either 0 or 1. They are the easiest type and are provided in benchmark attribute datasets AWA1, AWA2, CUB, SUN, aPY. Relative attributes \cite{parikh2011relative} on the other hand, show the strength of an attribute in a given image comparing to the other images. The real-valued attributes are in continuous form thus they have the best quality \cite{akata2015evaluation}. In the SUN attribute dataset \cite{patterson2012sun}, they have achieved confidence through averaging the binary labels from multiple annotators.

\subsubsection{Unsupervised Word-Embeddings}
Also known as Textual corpora embedding. Bag of Words (BOW) \cite{harris1954distributional} is a one-hot encoding approach. It simply shows the number of occurrences of the words in a representation called bag and is negligent of word orders and grammar.
One-hot encoding approaches had a drawback of giving the stop words (like "a", "the" and "of") high relevancy counts. Later,
Term Frequency-Inverse Document Frequency (TF-IDF): \cite{salton1988term} used term weighting to alleviate this problem by filtering the stop words and to keep meaningful words.
Word2Vec \cite{mikolov2013distributed}, a widely used two-layered neural embedding model and has two variants, CBOW and skip-gram. CBOW predicts a target word in the centre of a context using its surroundings words while the skip-gram model predicts surrounding words using a target word. CBOW is faster in train and usually results in better accuracy for frequent words while Skip-gram is preferred for rare words and it works well with sparse training data.
Global Vectors (GloVe) \cite {pennington2014glove} is trained on Wikipedia. It combines local context window methods and global matrix factorization. Glove learns to consider global word-word co-occurrence matrix statistics to build the word embeddings.

\subsubsection{Hierarchy Embedding}
WordNet \cite{miller1995wordnet} is a large-scale public lexical database of  117,000 synsets. Synsents are a group of words that are semantically related to each other. i.e. synonyms, homonyms and meronymies of English words that are organised using the hierarchy distances with a graph structure, thus Approaches based on knowledge graphs often follow the WordNet to measure the similarity between the word meanings \cite{rohrbach2010helps}, \cite{rohrbach2011evaluating}, \cite{akata2013label}, \cite{akata2015evaluation}, \cite{xian2016latent}, \cite{lu2015unsupervised}, \cite{akata2016multi}, \cite{qiao2016less}, \cite{marino2016more}, \cite{wang2018zero}, \cite{lee2018multi}.

\subsubsection{Language Modelling}
In the general ZSL scenarios, word by word representations considered; however, with the advent of transfer learning in the natural language processing (NLP), and the introduction of contextual word embeddings, the boundaries of the capabilities of the embeddings has been pushed further. Unlike the traditional word embeddings, language models can capture the meaning of the words based on the context in which they appear. Several contextual representations that have been introduced recently and showed great results. These existing pre-trained models can be fined tuned on various ZSL tasks. 

ELMo \cite{peters2018deep} is a contextual embedding model. Following  morphological clues together with a deep bidirectional language model (biLM), ELMo learns the representations. Bidirectional Encoder Representations from Transformer (BERT) \cite{devlin2018bert} is a multi-layer bidirectional Transformer encoder \cite{vaswani2017attention} trained upon BooksCorpus \cite{zhu2015aligning} dataset and English Wikipedia. It outperforms ELMo with having more parameters and layers.The pre-trained BERT model can be fine-tuned with just one additional output layer. However, BERT suffers from fine-tuning discrepancy due to ignoring the relation the masked positions have. XLNet \cite{yang2019xlnet} uses an autoregressive model to introduce a method that overcomes the shortcoming of BERT. In addition to the datasets used by BERT, XLNet pre-trains the model on Giga5 \cite{parker2011english}, ClueWeb 2012-B extended by \cite{callan2009clueweb09} and Common Crawl$^*$\footnote{$^*$http://commoncrawl.org}. ALBERT \cite{lan2019albert} increases the model size. It lowers the memory usage with two parameter reduction techniques. The first one is a factorized embedding parameterization. The second one is cross-layer parameter sharing. These two techniques result in lower memory usage and higher training speed than BERT. The data used for pre-training is the same as XLNeT. 

In this article, we report the results of ZSL and GZSL using the same class embeddings as \cite{xian2018zero} that is Word2Vec trained on Wikipedia for ImageNet and per-class attributes for the attribute datasets, and for the seen-unseen relatedness task we follow \cite{elhoseiny2017link} and consider TF-IDF for the CUB and NAB datasets.
\subsection{Image Embeddings}
\noindent Existing models use either shallow or deep feature representation. Examples of shallow features are SIFT \cite{lowe2004distinctive}, PHOG \cite{bosch2007representing}, SURF \cite{bay2008speeded} and local self-similarity histograms \cite{shechtman2007matching}. Among the mentioned features, SIFT is the commonly used features in ZSL models like \cite{akata2015label}, \cite{changpinyo2016synthesized} and \cite{fu2015transductive}.

Deep features are obtained from deep CNN architectures~\cite{mtei2019cnn} and contain higher-level features. Extracted features are one of the followings:\\

4,096-dim top-layer hidden unit activations (fc7) of the AlexNet \cite{krizhevsky2012imagenet},
1000-dim last fully connected layer (fc8) of VGG-16 \cite{simonyan2014very}, 
4,096-dim of the 6th layer (fc6) and
4,096-dim of the last layer (fc7) features of the VGG-19 \cite{simonyan2014very}. 1,024-dim top-layer pooling units of the GoogleNet \cite{szegedy2015going}. and
2048-dim last layer pooling units of the ResNet-101 \cite{he2016deep}.

In this paper, we consider the ResNet-101 network which is pre-trained on ImageNet-1K without any fine-tuning. That is the same image embedding used in \cite{xian2018zero}. Features are extracted from whole images of SUN, CUB, AWA1, AWA2, and ImageNet and the cropped bounding boxes of aPY. For the seen-unseen relatedness task, VGG-16 is used for CUB and NAB as proposed in \cite{elhoseiny2017link}.\\
%
\subsection{Evaluation Metrics}
\noindent Common evaluation criteria used for ZSL challenge are:\\

\textbf{Classification accuracy.} One of the simplest metrics is classification accuracy in which the ratio of the number of the correct predictions to samples in class $c$ is measured. However, it results in a bias towards the populated classes.\\

\textbf{Average per-class accuracy.} To reduce the bias problem for the populated classes, average per-class accuracies are computed by multiplying the division of the classification accuracy to division of their cumulative sum.
\begin{equation*}
acc_y=\frac{1}{\| y \|}\displaystyle \sum^{\| y \|}_{y=1}\frac{\# \mathrm{correct\, predictions\, in\, class\, } y}{\mathrm{\# samples\, in\, class\, } y} 
\quad \mbox{\cite{xian2017zero}}
\end{equation*}
\textbf{Harmonic mean.} For performance evaluation on both seen and unseen classes (i.e. the GZSL setting), the Top-1 accuracies for the seen and unseen classes are used to compute the harmonic mean:
\begin{equation*}
\quad H=\frac{2*acc_{y^S}*acc_{y^U}}{acc_{y^S}+acc_{y^U}} \quad \mbox{\cite{xian2017zero}}
\end{equation*}
In this paper, we designate the Top-1 accuracies and the harmonic mean as the evaluation protocols.\\
%
\section{Experimental Results \label{exper}}
\noindent As the main contributions of this research, and for the first time, we provide a comprehensive experiments of 21 state-of-the-art models in ZSL/GZSL domain that include the evaluations and comparisons of data-synthesising methods. 
In this section, first we provide the results for ZSL, GZSL and seen-unseen relatedness on attribute datasets, then we present the experimental results on the ImageNet dataset. A minor part of the results is reported from \cite{xian2017zero} for a more comprehensive comparison. 
\begin{table*}[t]
\caption{Zero-shot learning results for the Standard Split (SS) and Proposed Split (PS) on SUN, CUB, AWA1, AWA2, and aPY datasets. We measure Top-1 accuracy in $\%$ for the results. \textdagger and \textdaggerdbl denote inductive and transductive settings respectively\\} 
\hspace{1mm}
\footnotesize
\centering
    \begin{tabular}{ll|cc|cc|cc|cc|cc} 
      \toprule 
      \multirow{22}{0.3mm}{\textdagger} & \multirow{2}{*}{Methods} & \multicolumn{2}{c|}{\textbf{SUN}} & \multicolumn{2}{c|}{\textbf{CUB}} & \multicolumn{2}{c|}{\textbf{AWA1}} & \multicolumn{2}{c|}{\textbf{AWA2}} & \multicolumn{2}{c}{\textbf{aPY}}\\
      
      &  & SS & PS & SS & PS & SS & PS & SS & PS & SS & PS\\
      
      \midrule 
      &DAP \cite{lampert2013attribute}& 38.9 & 39.9 & 37.5 & 40.0 & 57.1 & 44.1 & 58.7 & 46.1 & 35.2 & 33.8\\
      
      &IAP \cite{lampert2013attribute}& 17.4 & 19.4 & 27.1 & 24.0 & 48.1 & 35.9 & 46.9 & 35.9 & 22.4 & 36.6\\
      
      &ConSE \cite{norouzi2013zero} & 44.2 & 38.8 & 36.7 & 34.3 & 63.6 & 45.6 & 67.9 & 44.5 & 25.9 & 26.9\\
      
      &CMT \cite{socher2013zero} & 41.9 & 39.9 & 37.3 & 34.6 & 58.9 & 39.5 & 66.3 & 37.9 & 26.9 & 28.0\\
      
      &SSE \cite{zhang2015zero} & 54.5 & 51.5 & 43.7 & 43.9 & 68.8 & 60.1 & 67.5 & 61.0 & 31.1 & 34.0\\
      
      &LATEM \cite{xian2016latent} & 56.9 & 55.3 & 49.4 & 49.3 & 74.8 & 55.1 & 68.7 & 55.8 & 34.5 & 35.2\\
      
      &ALE \cite{akata2015label} & 59.1 & 58.1 & 53.2 & 54.9 & 78.6 & 59.9 & 80.3 & 62.5 & 30.9 & 39.7\\
      
      &DeViSE \cite{frome2013devise} & 57.5 & 56.5 & 53.2 & 52.0 & 72.9 & 54.2 & 68.6 & 59.7 & 35.4 & 39.8\\
      
      &SJE \cite{akata2015evaluation} & 57.1 & 53.7 & 55.3 & 53.9 & 76.7 & 65.6 & 69.5 & 61.9 & 32.0 & 32.9\\
      
      &ESZSL \cite{romera2015embarrassingly} & 57.3 & 54.5 & 55.1 & 53.9 & 74.7 & 58.2 & 75.6 & 58.6 & 34.4 & 38.3\\
      
      &SYNC \cite{changpinyo2016synthesized} & 59.1 & 56.3 & 54.1 & 55.6 & 72.2 & 54.0 & 71.2 & 46.6 & 39.7 & 23.9\\
      
      &SAE \cite{kodirov2017semantic} & 42.4 & 40.3 & 33.4 & 33.3 & 80.6 & 53.0 & 80.7 & 54.1 & 8.3 & 8.3\\
      
      &GFZSL \cite{verma2017simple} & 62.9 & 60.6 & 53.0 & 49.3 & 80.5 & 68.3 & 79.3 & 63.8 & \textbf{51.3} & 38.4\\
      
      &DEM \cite{zhang2017learning} & - & 61.9 & - & 51.7 & - & 68.4 & - & 67.1 & - & 35.0\\
      
      &GAZSL \cite{zhu2018generative} & - & 61.3 & - & 55.8 & - & 68.2 & - & 68.4 & - & \textbf{41.1}\\
      
      &f-CLSWGAN \cite{xian2018feature}\,  & - & 60.8 & - & 57.3 & - & 68.8 & - & 68.2 & - & 40.5\\
      
      &CVAE-ZSL \cite{mishra2018generative} & - & 61.7 & - & 52.1 & - & \textbf{71.4} & - & 65.8 & - & -\\    
      
      &SE-ZSL \cite{kumar2018generalized} & \textbf{64.5} & \textbf{63.4} & \textbf{60.3} & \textbf{59.6} & \textbf{83.8} & 69.5 & \textbf{80.8} & \textbf{69.2} & - & -\\  
      
      \midrule      
      \multirow{2.6}{*}{\textdaggerdbl} &ALE-tran \cite{akata2015label} & - & 55.7 & - & \textbf{54.5} & - & 65.6 & - & 70.7 & - & \textbf{46.7}\\
      
      &GFZSL-tran \cite{verma2017simple} & - & \textbf{64.0} & - & 49.3 & - & \textbf{81.3} & - & \textbf{78.6} & - & 37.1\\
      
      &DSRL \cite{ye2017zero} & - & 56.8 & - & 48.7 & - & 74.7 & - & 72.8 & - & 45.5\\
      
      \bottomrule 
		\end{tabular}
  \label{tab:ssps}
\end{table*}

\subsection{Zero-Shot Learning Results}

\noindent For the original ZSL task where only unseen classes are being estimated during the test time, we compare 21 state-of-the-art models in Table~\ref{tab:ssps}, among which, DAP \cite{lampert2013attribute}, IAP \cite{lampert2013attribute} and ConSE \cite{norouzi2013zero} belong to attribute classifiers. CMT \cite{socher2013zero}, LATEM \cite{xian2016latent}, ALE \cite{akata2015label}, DeViSE \cite{frome2013devise}, SJE \cite{akata2015evaluation}, ESZSL \cite{romera2015embarrassingly}, GFZSL \cite{verma2017simple} and DSRL \cite{ye2017zero} are from compatibility learning approaches, SSE \cite{zhang2015zero} and SYNC \cite{changpinyo2016synthesized} are representative models of cross-modal embedding, DEM \cite{zhang2017learning}, GAZSL \cite{zhu2018generative}, f-CLSWGAN \cite{xian2018feature}, CVAE-ZSL \cite{mishra2018generative}, SE-ZSL \cite{kumar2018generalized} are visual embedding models. From the hybrid or combination category, we compare the results of SAE \cite{kodirov2017semantic}. Three transductive approaches ALE-tran \cite{akata2015label}, GFZSL-tran \cite{verma2017simple} and DSRL \cite{ye2017zero} are also presented among the selected models. Due to  the intrinsic nature of the transductive setting, the results are competitive and in some cases better than the inductive methods, i.e. for GFZSL-tran~\cite{verma2017simple} the accuracy is 9.9$\%$ higher than CVAE-ZSL \cite{mishra2018generative} for PS split of AWA1 dataset. However, in comparison with the inductive form of the same model, there are cases where the inductive model has better accuracies. i.e. in PS split of the aPY dataset, the performance is 38.4$\%$ vs 37.1$\%$ or for ALE-tran \cite{akata2015label} model in PS split of SUN it's 58.1$\%$ vs 55.7$\%$, also for PS split of CUB it is 54.9$\%$ vs 54.5$\%$ with its inductive type. GFZSL \cite{verma2017simple}, a compatibility-based approach, has the best scores compared to other models of the same category in every dataset except for the CUB where SJE \cite{akata2015evaluation} tops the results in both splits. This superiority could be due to the generative nature of the model. GFZSL \cite{verma2017simple} performs the best on AWA1 both in inductive and transductive settings. Out of cross-modal methods, SYNC \cite{changpinyo2016synthesized} performs better than SSE\cite{zhang2015zero} in SUN and CUB datasets, while for AWA1, AWA2 and aPY in SS split it has lower performance than SSE \cite{zhang2015zero} in the proposed split. Visual generative methods have proved to perform better as they make the problem into the traditional supervised form, among which, SE-ZSL \cite{kumar2018generalized} has the most outstanding performance. For the proposed split in one case on CUB dataset, SE-ZSL \cite{kumar2018generalized} performs better than ALE-tran \cite{akata2015label} which is its transductive counterpart where the accuracies are 59.6$\%$ vs 54.5$\%$. In PS split of AWA1, CVAE-ZSL \cite{mishra2018generative} stays at the top, with 1.9$\%$ higher accuracy than the second-best performing model. The accuracies for SS splits are higher than PS in most cases and the reason could be the test images included in training samples, especially for AWA1 and AWA2, as reported in \cite{xian2018zero}.

\begin{table*}[t]
\caption{Generalised Zero-Shot Learning results for the Proposed Split (PS) on SUN, CUB, AWA1, AWA2, and aPY datasets. We measure the Top-1 accuracy in $\%$ for seen ($y^S$), unseen ($y^U$) and their harmonic mean (H). \textdagger and \textdaggerdbl denote inductive and transductive settings, respectively.\\}
\centering
\footnotesize
\resizebox{1.0\textwidth}{!}{%
\begin{tabular}{ll|ccc|ccc|ccc|ccc|ccc} 
\toprule 
\multirow{24}{0.03mm}{\textdagger} & \multirow{2}{*}{Methods} & \multicolumn{3}{c|}{\textbf{SUN}} & \multicolumn{3}{c|}{\textbf{CUB}} & \multicolumn{3}{c|}{\textbf{AWA1}} & \multicolumn{3}{c|}{\textbf{AWA2}} & \multicolumn{3}{c}{\textbf{aPY}}\\
& & $y^U$ & $y^S$ & H & $y^U$ & $y^S$ & H & $y^U$ & $y^S$  & H& $y^U$ & $y^S$ & H & $y^U$ & $y^S$ & H\\
\midrule 
& DAP \cite{lampert2013attribute}& 4.2 & 25.1 & 7.2 & 1.7 & 67.9 & 3.3 & 0.0 & \textbf{88.7} & 0.0 & 0.0 & 84.7 & 0.0 & 4.8 & 78.3 & 9.0\\
& IAP \cite{lampert2013attribute}& 1.0 & 37.8 & 1.8 & 0.2 & \textbf{72.8} & 0.4 & 2.1 & 78.2 & 4.1 & 0.9 & 87.6 & 1.8 & 5.7 & 65.6 & 10.4\\
& ConSE \cite{norouzi2013zero}  & 6.8 & 39.9 & 11.6 & 1.6 & 72.2 & 3.1 & 0.4 & 88.6 & 0.8 & 0.5 & \textbf{90.6} & 1.0 & 0.0 & \textbf{91.2} & 0.0\\
& CMT \cite{socher2013zero} & 8.1 & 21.8 & 11.8 & 7.2 & 49.8 & 12.6 & 0.9 & 87.6 & 1.8 & 0.5 & 90.0 & 1.0 & 1.4 & 85.2 & 2.8\\
& CMT* \cite{socher2013zero} & 8.7 & 28.0 & 13.3 & 4.7 & 60.1 & 8.7 & 8.4 & 86.9 & 15.3 & 8.7 & 89.0 & 15.9 & 10.9 & 74.2 & 19.0\\      
& SSE \cite{zhang2015zero} & 2.1 & 36.4 & 4.0 & 8.5 & 46.9 & 14.4 & 7.0 & 80.5 & 12.9 & 8.1 & 82.5 & 14.8 & 0.2 & 78.9 & 0.4\\
& LATEM \cite{xian2016latent} & 14.7 & 28.8 & 19.5 & 15.2 & 57.3 & 24.0 & 7.3 & 71.7 & 13.3 & 11.5 & 77.3 & 20.0 & 0.1 & 73.0 & 0.2\\
& ALE \cite{akata2015label} & 21.8 & 33.1 & 26.3 & 23.7 & 62.8 & 34.4 & 16.8 & 76.1 & 27.5 & 14.0 & 81.8 & 23.9 & 4.6 & 73.7 & 8.7\\
& DeViSE \cite{frome2013devise} & 16.9 & 27.4 & 20.9 & 23.8 & 53.0 & 32.8 & 13.4 & 68.7 & 22.4 & 17.1 & 74.7 & 27.8 & 4.9 & 76.9 & 9.2\\
& SJE \cite{akata2015evaluation} & 14.7 & 30.5 & 19.8 & 23.5 & 59.2 & 33.6 & 11.3 & 74.6 & 19.6 & 8.0 & 73.9 & 14.4 & 3.7 & 55.7 & 6.9\\
& ESZSL \cite{romera2015embarrassingly} & 11.0 & 27.9 & 15.8 & 12.6 & 63.8 & 21.0 & 6.6 & 75.6 & 12.1 & 5.9 & 77.8 & 11.0 & 2.4 & 70.1 & 4.6\\  
& SYNC \cite{changpinyo2016synthesized} & 7.9 & \textbf{43.3} & 13.4 & 11.5 & 70.9 & 19.8 & 8.9 & 87.3 & 16.2 & 10.0 & 90.5 & 18.0 & 7.4 & 66.3 & 13.3\\
& SAE \cite{kodirov2017semantic} & 8.8 & 18.0 & 11.8 & 7.8 & 54.0 & 13.6 & 1.8 & 77.1 & 3.5 & 1.1 & 82.2 & 2.2 & 0.4 & 80.9 & 0.9\\
& GFZSL \cite{verma2017simple} & 0.0 & 39.6 & 0.0 & 0.0 & 45.7 & 0.0 & 1.8 & 80.3 & 3.5 & 2.5 & 80.1 & 4.8 & 0.0 & 83.3 & 0.0\\
& DEM \cite{zhang2017learning} & 20.5 & 34.3 & 25.6 & 19.6 & 57.9 & 29.2 & 32.8 & 84.7 & 47.3 & 30.5 & 86.4 & 45.1 & 11.1 & 75.1 & 19.4\\      
& GAZSL \cite{zhu2018generative} & 21.7 & 34.5 & 26.7 & 23.9 & 60.6 & 34.3 & 25.7 & 82.0 & 39.2 & 19.2 & 86.5 & 31.4 & 14.2 & 78.6 & 24.1\\
& f-CLSWGAN \cite{xian2018feature}\, & 42.6 & 36.6 & 39.4 & 43.7 & 57.7 & 49.7 & \textbf{57.9} & 61.4 & 59.6 & 52.1 & 68.9 & 59.4 & \textbf{32.9} & 61.7 & \textbf{42.9}\\          
& CVAE-ZSL \cite{mishra2018generative} & - & - & 26.7 & - & - & 34.5 & - & - & 47.2 & - & - & 51.2 & - & - & -\\
& SE-GZSL \cite{kumar2018generalized} & 40.9 & 30.5 & 34.9 & 41.5 & 53.3 & 46.7 & 56.3 & 67.8 & 61.5 & \textbf{58.3} & 68.1 & 62.8 & - & - & -\\    
& CADA-VAE \cite{schonfeld2019generalized} & \textbf{47.2} & 35.7 & \textbf{40.6} & \textbf{51.6} & 53.5 & \textbf{52.4} & 57.3 & 72.8 & \textbf{64.1} & 55.8 & 75.0 & \textbf{63.9} & - & - & -\\
\midrule
\multirow{3}{*}{\textdaggerdbl} & ALE-tran \cite{akata2015label} & \textbf{19.9} & 22.6 & \textbf{21.2} & 23.5 & 45.1 & 30.9 & 25.9 & - & - & 12.6 & 73.0 & 21.5 & 8.1 & - & -\\
& GFZSL-tran \cite{verma2017simple} & 0 & \textbf{41.6} & 0 & \textbf{24.9} & \textbf{45.8} & \textbf{32.2} & \textbf{48.1} & - & - & \textbf{31.7} & 67.2 & \textbf{43.1} & 0.0 & - & -\\
& DSRL \cite{ye2017zero} & 17.7 & 25.0 & 20.7 & 17.3 & 39.0 & 24.0 & 22.3 & - & - & 20.8 & \textbf{74.7} & 32.6 & \textbf{11.9} & - & -\\
\bottomrule  
  \end{tabular}}
  \label{tab:gzslsuh}
\end{table*}
%

\subsection{Generalised Zero-Shot Learning Results}

\noindent A more real-world scenario where previously learned concepts are estimated alongside new ones is necessary to experiment. 21 state-of-the-art models, same as ZSL challenge, include: DAP \cite{lampert2013attribute}, IAP  \cite{lampert2013attribute}, ConSE \cite{norouzi2013zero}, CMT \cite{socher2013zero}, SSE \cite{zhang2015zero}, LATEM \cite{xian2016latent}, ALE \cite{akata2015label}, DeViSE \cite{frome2013devise}, SJE \cite{akata2015evaluation} ,ESZSL \cite{romera2015embarrassingly}, SYNC \cite{changpinyo2016synthesized}, SAE \cite{kodirov2017semantic}, GFZSL \cite{verma2017simple}, DEM \cite{zhang2017learning}, GAZSL \cite{zhu2018generative}, f-CLSWGAN \cite{xian2018feature}, CVAE-ZSL \cite{mishra2018generative}, SE-GZSL \cite{kumar2018generalized}, ALE-train \cite{akata2015label}, GFZSL-tran \cite{verma2017simple}, DSRL \cite{ye2017zero}. CADA-VAE \cite{schonfeld2019generalized} is added to the comparison as a model combining the visual feature augmentation approach with the cross-modal alignment. CMT* \cite{socher2013zero} has a novelty detection and is included in the report as an alternative version to CMT \cite{socher2013zero}. The reports in Table~\ref{tab:gzslsuh} are in PS splits. As shown in the table, the results on $y^{S}$ are dramatically higher than $y^{U}$ since in GZSL, the test search space includes seen classes as well as unseen classes,  this gap is the most conspicuous in attribute classifiers like DAP  \cite{lampert2013attribute} that performs poorly on AWA1 and AWA2, hybrid approaches and in GFZSL \cite{verma2017simple} where it results in 0$\%$ accuracy on SUN and CUB when training classes are estimated at test time. However for three models f-CLSWGAN \cite{xian2018feature}, SE-GZSL \cite{kumar2018generalized} and CADA-VAE \cite{schonfeld2019generalized} in SUN dataset, the accuracy for $y^U$ is higher than $y^{S}$, i.e. for SE-GZSL \cite{kumar2018generalized} it is 10.4$\%$ higher. For a fair comparison, the weighted average of training and test classes is also reported. According to harmonic means, the best model on all evaluated datasets is SE-ZSL \cite{kumar2018generalized}, although the results haven't been reported for aPY. In some cases, the attribute classifier achieves the best results on $y^{S}$. Transductive models have fluctuating results in comparison with their inductive types. CADA-VAE \cite{schonfeld2019generalized} achieves the best performance in all of the harmonic means cases (results for aPY are not reported) and shows the best results, higher than all of the transductive methods.

\begin{figure*}
\centering
\subfloat[Zero-Shot Learning]
{
\begin{tikzpicture}    
    \begin{axis}[
        width  = 0.505\textwidth,
        height = 0.3\textheight,
        compat=1.8,
        major x tick style = transparent,
        ybar=1.6*\pgflinewidth,
        bar width=1.3pt,
        ymajorgrids = true,
        xmajorgrids = true,
        label style={font=\small},
        ylabel = {Top-1 Acc. (in \%)},
        symbolic x coords={2H,3H,M500,M1K,M5K,L500,L1K,L5K,All},
        xtick= {2H,3H,M500,M1K,M5K,L500,L1K,L5K,All},
        xtick = data,
        scaled x ticks = true,
        scaled y ticks = true,
        enlarge x limits=0.075,
        ymin=0,     
        legend cell align=left,
        legend style={at={(1,1)},anchor=north east}
    ]
        \addplot[style={ao,fill=ao,mark=none,draw=ao!80!black}]
             coordinates {(2H,7.63) (3H,2.18) (M500,12.33) (M1K,8.31) (M5K,3.22) (L500,3.53) (L1K,2.69) (L5K,1.05) (All,0.95)};

        \addplot[style={spirodiscoball,fill=spirodiscoball,mark=none,draw=spirodiscoball!80!black}]
             coordinates {(2H,2.88) (3H,0.67) (M500,5.10) (M1K,3.04) (M5K,1.04) (L500,1.87) (L1K,1.08) (L5K,0.33) (All,0.29)};              

        \addplot[style={azurecw,fill=azurecw,mark=none,draw=azurecw!80!black}]
             coordinates {(2H,5.45) (3H,1.32) (M500,10.81) (M1K,6.63) (M5K,1.90) (L500,4.53) (L1K,2.74) (L5K,0.76) (All,0.50)}; 

        \addplot[style={aqua,fill=aqua,mark=none,draw=aqua!80!black}]
             coordinates {(2H,5.38) (3H,1.32) (M500,10.40) (M1K,6.77) (M5K,2.00) (L500,4.27) (L1K,2.85) (L5K,0.79) (All,0.50)}; 

        \addplot[style={guppiegreen,fill=guppiegreen,mark=none,draw=guppiegreen!80!black}]
             coordinates {(2H,5.25) (3H,1.29) (M500,10.36) (M1K,6.68) (M5K,1.94) (L500,4.23) (L1K,2.86) (L5K,0.78) (All,0.49)};

        \addplot[style={ufo,fill=ufo,mark=none,draw=ufo!80!black}]
             coordinates {(2H,5.31) (3H,1.33) (M500,9.88) (M1K,6.53) (M5K,1.99) (L500,4.93) (L1K,2.93) (L5K,0.78) (All,0.52)};

        \addplot[style={yellow,fill=yellow,mark=none,draw=yellow!80!black}]
             coordinates {(2H,6.35) (3H,1.51) (M500,11.91) (M1K,7.69) (M5K,2.34) (L500,4.50) (L1K,3.23) (L5K,0.94) (All,0.64)};

        \addplot[style={amber,fill=amber,mark=none,draw=amber!80!black}]
             coordinates {(2H,9.26) (3H,2.29) (M500,15.83) (M1K,10.75) (M5K,3.42) (L500,5.83) (L1K,3.52) (L5K,1.26) (All,0.96)};     

        \addplot[style={safetyorange,fill=safetyorange,mark=none,draw=safetyorange!80!black}]
            coordinates {(2H,4.89) (3H,1.26) (M500,9.96) (M1K,6.57) (M5K,2.09) (L500,2.5) (L1K,2.17) (L5K,0.72) (All,0.56)};

        \addplot[style={tractorred,fill=tractorred,mark=none,draw=tractorred!80!black}]
             coordinates {(2H,10.0) (3H,2.3) (M500,15.8) (M1K,10.9) (M5K,3.3) (L500,6.0) (L1K,3.62) (L5K,1.10) (All,1.00)};
                         
        \addplot[style={bostonuniversityred,fill=bostonuniversityred,mark=none,draw=bostonuniversityred!80!black}]
             coordinates {(2H,0.0) (3H,0.0) (M500,0.0) (M1K,0.0) (M5K,0.0) (L500,0.0) (L1K,0.0) (L5K,0.0) (All,0.0)};                 
                                                                                                                        
        \addplot[style={darkmagneta,fill=darkmagneta,mark=none,draw=darkmagneta!80!black}]
             coordinates {(2H,13.22) (3H,3.3) (M500,0.0) (M1K,0.0) (M5K,0.0) (L500,0.0) (L1K,0.0) (L5K,0.0) (All,1.80)}; 
                                                                                                                     
        \legend{ConSE,CMT,LATEM,ALE,DeViSE,SJE,ESZSL,SYNC,SAE,f-CLSWGAN,CADA-VAE,f-VAEGAN-D2}
    \end{axis}  
\end{tikzpicture}
\label{fig:imagenetgzsl}  
}\hspace{5mm}
\subfloat[Generalised Zero-Shot Learning]
{
\begin{tikzpicture}    
    \begin{axis}[
        width  = 0.505\textwidth,
        height = 0.3\textheight,
        compat=1.8,
        major x tick style = transparent,
        ybar=1.6*\pgflinewidth,
        bar width=1.3pt,
        ymajorgrids = true,
        xmajorgrids = true,
        label style={font=\small},
        ylabel = {Top-1 Acc. (in \%)},
        symbolic x coords={2H,3H,M500,M1K,M5K,L500,L1K,L5K,All},
        xtick= {2H,3H,M500,M1K,M5K,L500,L1K,L5K,All},
        xtick = data,
        scaled x ticks = true,
        scaled y ticks = true,
        enlarge x limits=0.075,
        ymin=0,     
        legend cell align=left,
        legend style={at={(1,1)},anchor=north east}
    ]
        \addplot[style={ao,fill=ao,mark=none,draw=ao!80!black}]
             coordinates {(2H,0.0) (3H,0.01) (M500,0.04) (M1K,0.038) (M5K,0.036) (L500,0.0) (L1K,0.0) (L5K,0.0) (All,0.0)};

        \addplot[style={spirodiscoball,fill=spirodiscoball,mark=none,draw=spirodiscoball!80!black}]
             coordinates {(2H,1.1) (3H,0.3) (M500,1.57) (M1K,1.27) (M5K,0.56) (L500,0.77) (L1K,0.4) (L5K,0.2) (All,0.21)};              

        \addplot[style={azurecw,fill=azurecw,mark=none,draw=azurecw!80!black}]
             coordinates {(2H,2.0) (3H,0.75) (M500,2.65) (M1K,2.08) (M5K,1.02) (L500,1.03) (L1K,0.53) (L5K,0.38) (All,0.37)}; 

        \addplot[style={aqua,fill=aqua,mark=none,draw=aqua!80!black}]
             coordinates {(2H,2.18) (3H,0.77) (M500,2.88) (M1K,2.32) (M5K,1.09) (L500,1.83) (L1K,1.22) (L5K,0.43) (All,0.32)}; 

        \addplot[style={guppiegreen,fill=guppiegreen,mark=none,draw=guppiegreen!80!black}]
             coordinates {(2H,2.15) (3H,0.76) (M500,2.90) (M1K,2.31) (M5K,1.08) (L500,1.61) (L1K,1.27) (L5K,0.43) (All,0.31)};

        \addplot[style={ufo,fill=ufo,mark=none,draw=ufo!80!black}]
             coordinates {(2H,1.77) (3H,0.72) (M500,2.32) (M1K,1.81) (M5K,0.92) (L500,1.63) (L1K,1.32) (L5K,0.42) (All,0.29)};

        \addplot[style={yellow,fill=yellow,mark=none,draw=yellow!80!black}]
             coordinates {(2H,1.29) (3H,0.53) (M500,1.69) (M1K,1.5) (M5K,0.76) (L500,0.66) (L1K,0.42) (L5K,0.27) (All,0.27)};

        \addplot[style={amber,fill=amber,mark=none,draw=amber!80!black}]
             coordinates {(2H,0.0) (3H,0.0) (M500,0.0) (M1K,0.0) (M5K,0.0) (L500,0.0) (L1K,0.0) (L5K,0.04) (All,0.0)};     

        \addplot[style={safetyorange,fill=safetyorange,mark=none,draw=safetyorange!80!black}]
            coordinates {(2H,0.77) (3H,0.28) (M500,1.68) (M1K,1.27) (M5K,0.53) (L500,0.06) (L1K,0.055) (L5K,0.0) (All,0.19)};

        \addplot[style={tractorred,fill=tractorred,mark=none,draw=tractorred!80!black}]
             coordinates {(2H,4.3) (3H,1.4) (M500,4.7) (M1K,4.1) (M5K,1.85) (L500,3.2) (L1K,2.5) (L5K,0.7) (All,0.40)};
                         
        \addplot[style={bostonuniversityred,fill=bostonuniversityred,mark=none,draw=bostonuniversityred!80!black}]
             coordinates {(2H,5) (3H,1.70) (M500,9.7) (M1K,6.25) (M5K,2.5) (L500,3.3) (L1K,3.1) (L5K,0.75) (All,0.55)};                 
                                                                                                                        
        \addplot[style={darkmagneta,fill=darkmagneta,mark=none,draw=darkmagneta!80!black}]
             coordinates {(2H,5.1) (3H,1.9) (M500,0.0) (M1K,0.0) (M5K,0.0) (L500,0.0) (L1K,0.0) (L5K,0.0) (All,0.95)}; 
                                                                                                                     
        \legend{ConSE,CMT,LATEM,ALE,DeViSE,SJE,ESZSL,SYNC,SAE,f-CLSWGAN,CADA-VAE,f-VAEGAN-D2}
    \end{axis}
\end{tikzpicture}
}
\caption{ImageNet results measured with Top-1 accuracy in $\%$ for the 9 splits including 2 and 3 hops away from ImageNet-1K training classes (2H and 3H) and 500, 1K and 5K most (M) and least (L) populated classes, and All the remaining ImageNet-20K classes.}
\label{fig4}
\end{figure*}
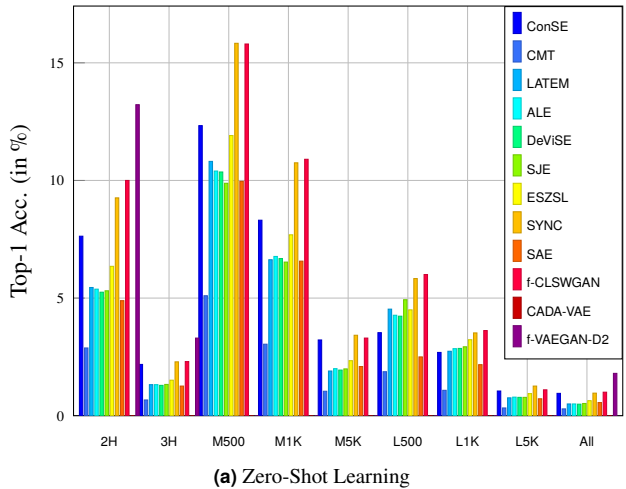
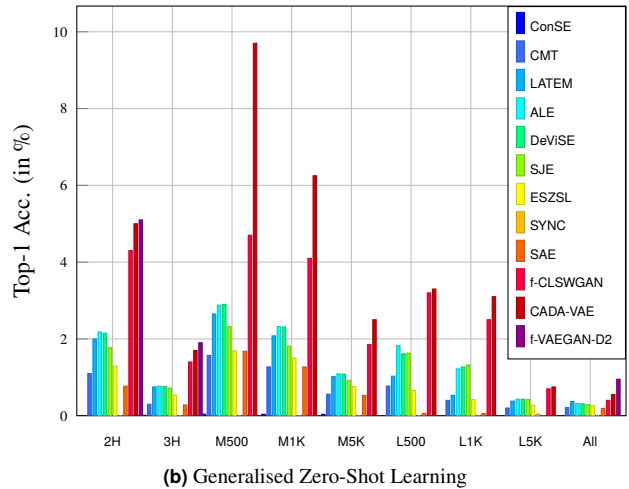

\subsection{Seen-Unseen Relatedness Results}
\noindent For fine-grained problems, sometimes it is important to measure the closeness of previously known concepts to novel unknown ones. For this purpose, a total of eleven models are compared in Table~\ref{tab:scssce}. MCZS \cite{akata2016multi}, WAC-Linear \cite{elhoseiny2013write}, WAC-Kernel \cite{elhoseiny2016write}, ESZSL \cite{romera2015embarrassingly}, SJE \cite{akata2015evaluation}, ZSLNS \cite{qiao2016less}, SynC$_{fast}$ \cite{changpinyo2016synthesized}, SynC$_{OVO}$ \cite{changpinyo2016synthesized}, ZSLPP \cite{elhoseiny2017link}, GAZSL \cite{zhu2018generative} and CANZSL \cite{chen2019canzsl}. SCE is the hard split thus has lower results compared to the SCS splits. The two variations reported for SYNC \cite{changpinyo2016synthesized} model, SynC$_{fast}$ denotes the setting in which the standard Crammer-Singer loss is used, and SynC$_{fast}$ \cite{changpinyo2016synthesized} depicts setting with one-versus-other classifiers. The first setting has better accuracies on CUB. CANZSL \cite{chen2019canzsl} outperforms all other models in both datasets and splits and improves the accuracy by 4$\%$ from 10.3$\%$ to 14.3$\%$ on SCE split of the CUB dataset and 35.6$\%$ vs 38.1$\%$ in SCS splits of NAB compared to the next best performing model is GAZSL \cite{zhu2018generative}. Similar to previous experiments, in the seen-unseen relatedness challenge, models that contain feature generating steps have the highest results.

\begin{table} [hbt!]
\caption{Seen-Unseen relatedness results on CUB and NAB datasets with easy (SCS) and hard (SCE) splits. Top-1 accuracy is reported in $\%$\\}
\centering
\footnotesize
    \begin{tabular}{l|cc|cc} 
      \toprule 
      & \multicolumn{2}{c|}{\textbf{CUB}} & \multicolumn{2}{c}{\textbf{NAB}}\\
       \multicolumn{1}{c|}{Methods} & SCS & SCE & SCS & SCE\\
      \midrule 
      MCZSL \cite{akata2016multi}  & 34.7 & - & - & -\\
      WAC-Linear \cite{elhoseiny2013write}\,  & 27.0 & 5.0 & - & -\\
      WAC-Kernel \cite{elhoseiny2016write}\,  & 33.5 & 7.7 & 11.4 & 6.0\\            
      ESZSL \cite{romera2015embarrassingly} & 28.5 & 7.4 & 24.3 & 6.3\\
      SJE \cite{akata2015evaluation}  & 29.9 & - & - & -\\
      ZSLNS \cite{qiao2016less} & 29.1 & 7.3 & 24.5 & 6.8\\
      SynC$_{fast}$ \cite{changpinyo2016synthesized} & 28.0 & 8.6 & 18.4 & 3.8\\
      SynC$_{OVO}$ \cite{changpinyo2016synthesized} & 12.5 & 5.9 & - & -\\
      ZSLPP \cite{elhoseiny2017link}  & 37.2 & 9.7 & 30.3 & 8.1\\  
      GAZSL \cite{zhu2018generative} & 43.7 & 10.3 & 35.6 & 8.6\\                    
      CANZSL \cite{chen2019canzsl} & \textbf{45.8} & \textbf{14.3} & \textbf{38.1} & \textbf{8.9}\\
      \bottomrule 
\end{tabular}
\label{tab:scssce}
\end{table}

\subsection{Zero-Shot Learning Results on ImageNet}
\label{zsl}
\noindent ImageNet is a large-scale single-labelled dataset with an imbalanced number of data that possesses WordNet hierarchy instead of human-annotated attributes, thus is useful mean to measure the performance of various methods in recognition-in-the-wild scenarios. The performances of 12 state-of-the-art models are reported here. They are ConSE \cite{norouzi2013zero}, CMT \cite{socher2013zero}, LATEM \cite{xian2016latent}, ALE \cite{akata2015label}, DeViSE \cite{frome2013devise}, SJE \cite{akata2015evaluation} ,ESZSL \cite{romera2015embarrassingly}, SYNC \cite{changpinyo2016synthesized}, SAE \cite{kodirov2017semantic}, f-CLSWGAN \cite{xian2018feature}, CADA-VAE \cite{schonfeld2019generalized} and f-VAEGAN-D2 \cite{xian2019f}. All of the Top-1 accuracies, except for the data generating models are reported from \cite{xian2018zero} experiments. As it can be understood from Figure~\ref{fig:imagenetgzsl}, Feature generating methods have outstanding performance compared to other approaches. Although the results of f-VAEGAN-D2 \cite{xian2019f} are available only for 2H, 3H and all splits, it still has the highest accuracies among other models. 
SYNC \cite{changpinyo2016synthesized} and f-CLSWGAN \cite{xian2018feature} are the next best performing models with approximately the same accuracies. ConSE \cite{norouzi2013zero} is a representative model from attribute-classifier based models, as it is also superior to direct compatibility approaches. ESZSL \cite{romera2015embarrassingly}, a model with linear compatibility function outperforms the other model within its category. However, in one case, SJE \cite{akata2015evaluation} has slightly better accuracy in L500 split setting.
It can be interpreted from the figures that on coarse-grained classes, the results are conspicuously better, while fine-grained classes with few images per class have more challenges. However, if the test search space is too big then the accuracies decrease. i.e. M5K has lower accuracies compared to L500 splits, and on 20K split, it is the lowest.

The GZSL results are important in the way that they depict the models' ability to recognise both seen and unseen classes at the test time. The results for the SYNC \cite{changpinyo2016synthesized} model is only reported in the L5K setting. As shown in Figure~\ref{fig:imagenetgzsl}, the trend is Similar to ZSL where populated classes have better results than the least populated classes, yet have poor results if the search spaces become too big like the decreasing trends in most and least populated classes. Moreover, data-generating approaches dominate other strategies. CADA-VAE \cite{schonfeld2019generalized} that has the advantages of both cross-modal alignment and data feature synthesising methods, evidently outperforms other models. In one case, i.e M500, it nearly has double the accuracy of f-CLSWGAN \cite{xian2018feature}. For the semantic embedding category, although ESZSL \cite{romera2015embarrassingly} had better results on ZSL, it falls behind approaches like ALE \cite{akata2015label}, DeViSE \cite{frome2013devise} and SJE \cite{akata2015evaluation}.

\section{Applications \label{App}}
\noindent During the very recent years, zero-shot learning has proved to be a necessary challenge to-be-solved for different scenarios and applications. The number of demands for learning without accessing to the unseen target concepts is also increasing each year.\\ 

Zero-shot learning is widely discussed in the computer vision field, such as object recognition in general, as in 
\cite{rezaei2015} and \cite{Sabzevari2008} where they aim to locate the objects beside recognising them. Several other variations of ZSL models are proposed for the same task purpose such as \cite{bansal2018zero}, \cite{rahman2018zero} and \cite{demirel2018zero}.
Zero-shot emotion recognition \cite{zhan2019zero} has the task of recognising unseen emotions while zero-shot semantic segmentation aims to segment the unseen object categories \cite{bucher2019zero} and \cite{wang2019zero}. Moreover, on the task of retrieving images from a large scale set of data, Zero-shot has a growing number of research \cite{long2018towards} \cite{xu2017attribute} along with sketch-based image retrieval systems 
\cite{dutta2019semantically}, \cite{dey2019doodle} and \cite{shen2018zero}. 
Zero-shot learning has an application on visual imitation learning to reduce human supervision by automating the exploration of the agent \cite{pathak2018zero}, \cite{lazaro2019beyond}.
Action recognition is the task of recognising the sequence of actions from the frames of a video. However, if the new actions are not available when training, Zero-shot learning can be a solution, such as in \cite{gao2019know}, \cite{qin2017zero}, \cite{mishra2018generativeaction} and \cite{shen2018scaling}.
Zero-shot Style Transfer in an image is the problem of transferring the texture of source image to target image while the style is not pre-determined and it is arbitrary \cite{sheng2018avatar}.
Zero-shot resolution enhancement problem aims at enhancing the resolution of an image without pre-defined high-resolution images for training examples \cite{shocher2018zero}.
Zero-shot scene classification for HSR images \cite{li2017zeroscene} and scene-sketch classification has been studied in \cite{xie2019deep} as other applications of ZSL in computer vision.\\
Zero-shot learning has also left its footprint in the area of NLP. Zero-Shot Entity Linking, links entity mentions in the text using a knowledge base \cite{logeswaran2019zero}. Many research works focus on the task of translating languages to another without pre-determined translation between pairs of samples~\cite{gu2019improved}, \cite{johnson2017google}, \cite{ha2017effective}, \cite{lakew2018improving}. In sentence embedding \cite{artetxe2019massively} and in Style transfer of text, a common technique is to convert the source to another style via arbitrary styles like the artistic technique discussed in \cite{carlson2017zero}.\\
In the audio processing field, zero-shot based voice conversion to another speaker's voice \cite{qian2019autovc} is an applicable scenario of ZSL.

In the era of the COVID-19 pandemic, many researchers have tried to work on Artificial Intelligence and Machine learning based methodologies to recognise the positive cases of the COVID-19 patients based on the CT scan images or Chest X-rays.
Two prominent features in chest CT used for diagnosis are ground glass opacities (GGO) and consolidation which has been considered by some of the researchers such as \cite{fang2020sensitivity}, \cite{ye2020chest}, \cite{li2020coronavirus}, and \cite{shan2020lung}. 
\cite{narin2020automatic} uses three CNN model to detect COVID-19, in which the ResNet50 shows a very high rate of classification performance. \cite{shan2020lung} introduces a deep-learning based system that segments the infected regions and the entire lung in an automatic manner. \cite{xia2020clinical} shows that the increase in unilateral or bilateral Procalcitonin and consolidation with surrounding halo is prominent in chest CT of paediatric patients. \cite{li2020artificial} introduces the COVNet to extract the 2D local and 3D global features in 3D chest CT slices. The method claims the ability of classifying COVID-19 from community acquired pneumonia (CAP). \cite{shi2020radiological} shows different imaging patterns of the COVID-19 cases depending on the time of infection. \cite{zheng2020time} classifies four stages to respiratory CT scan changes and shows the most dramatic changes to be in the first 10 days from the onset of initial symptoms.
\cite{zhang2020covid} introduce a deep learning based anomaly detection model which extracts the high-level features from the input chest X-ray image. \cite{hemdan2020covidx} introduce COVIDX-Net to classify the positive cases for the COVID-19 in X-ray images. It uses 7 different architectures, which VGG19 outperforms the others. \cite{afshar2020covid} propose a COVID-CAPS that is based on the Capsule Networks \cite{hinton2018matrix} to avoid the drawbacks of CNN-based architectures as it captures better spatial information. It performs on a small dataset of X-ray images. \cite{abbas2020classification} employ a class decomposition mechanism in DeTraC \cite{abbas2020detrac} which is a deep convolutional network that can handle image dataset irregularities of the X-ray images.
Zhang et al. \cite{zhang2020a} propose a method for X-ray medical image segmentation using task driven generative adversarial networks.
\cite{rajpurkar2017chexnet} proposes a 21-layer CNN called CheXNet, trained on the  ChestX-ray14 dataset \cite{wang2017chestx} to detect pneumonia with the localisation of the most infected areas from the X-ray images. \cite{rutigliano2019chronic} shows a possible diagnostic criteria could be the existence of  bilateral pulmonary  areas  of  consolidation found in the chest X-rays, and \cite{varshni2019pneumonia} use DenseNet-169 for the purpose of feature extraction followed by an SVM classifier to detect Pneumonia from chest X-ray images.

A common weakness among the majority of the above-mentioned research works is that they either conduct their evaluations on a very limited number of cases due to the lack of comprehensive datasets (which puts the validity of the reported results under a question),
or they suffer from underlying uncertainties due to unknown nature and characteristics of the novel COVID-19, not only for the medical community, but also for the machine learning and data analytic experts. 
In such an uncertain atmosphere with limited training dataset, we strongly recommend the adaptation of Zero-shot learning and its variances (as discussed in Figure \ref{fig4}) as an efficient deep learning based solution towards COVID-19 diagnosis.

Diagnosis and recognition of the very recent and global challenge of COVID-19 disease caused by the severe acute respiratory syndrome Coronavirus 2 (SARS-CoV-2) is a perfect real-world application of Zero-shot learning, where we do not have millions of annotated datasets available; and the symptoms of the disease and the chest X-ray of infected people may significantly vary from person to person. Such a scenario can truly be considered as a novel unseen target or classification challenge. We only know some of the symptoms of the infected people with COVID-19 in forms of advices, text notes, chest X-ray interpenetration, all as the auxiliary data which have partial similarities with other lung inflammatory diseases, such as Asthma or SARS. So, we have to seek for a semantic relationship between training and the new unseen classes. Therefore, ZSL can help us significantly to cope with this new challenge like the induction of the SARS-CoV-2, from previously learned diagnosis of the Asthma, and the Pneumonia using written medical documents of the respiratory tracts and chest X-ray images. In the case of the few-shot learning, a handful of the chest CT scans or X-ray of the positive cases of the COVID-19 can also be beneficial as further support-set alongside the chest X-ray images of SARS, Asthma and Pneumonia to infer the novel COVID-19 cases.

As a general rule and based on the recent successful applications, we can infer that in any scenarios that the goal is set to reduce supervision, and the target of the problem can be learned through side information and its relation to the seen data, the Zero-shot learning method can be conducted as one of the best learning techniques and practices.

\section{Discussion \label{discuss}}
\noindent A typical zero-shot learning problem is usually faced with three popular issues that need to be solved in order to enhance the performance of the model. These issues are Bias, Hubness and domain shift; and every model revolves around solving one or more of the issues mentioned. In this section, we discuss efforts done by different approaches to alleviate bias, hubness and domain-shift and infer the logic each approach owns to learn its model.\\

\textbf{Bias.} The problem with ZSL and GZSL tasks is that the imbalanced data between training and test classes cause a bias towards seen classes at prediction time. Other reasons for bias could be high-dimensionality and the devoid of manifold structure of features. Several data generating approaches have worked on alleviating bias by synthesising visual data for unseen classes. \cite{xian2018feature} generate semantically rich CNN features of the unseen classes to make unseen embedding space more known. \cite{mishra2018generative} generates pseudo seen and unseen class features, and then it trains an SVM classifier to mitigate bias. \cite{sariyildiz2019gradient} improve the quality of the synthesised examples by using gradient matching loss. Models combining data generation or reconstruction along with other techniques have proved to be effective in alleviating bias. \cite{long2017zero} use an intermediate space to help discover the geometric structure of the features that previously didn't with the regression-based projections. \cite{chen2018zero} used calibrated stacking rule. \cite{schonfeld2019generalized} generated latent feature sizes of 64 with the idea that low-dimensional representations tend to mitigate bias. \cite{shibing2019bi} uses two regressors to calculate reconstruction to diminish the bias. Transductive-based approaches like \cite{sariyildiz2019gradient} are also used to solve the bias issue. In \cite{song2018transductive}, it forces the unseen classes to be projected into fixed pre-defined points to avoid results with bias.\\

\textbf{Hubness} \cite{radovanovic2010hubs}. In large-dimensional mapping spaces, samples (hubs) might end up falsely as the nearest neighbours of several other points in the semantic space and result in an incorrect prediction. To avoid the hubness, \cite{wang2017zero} propose a stage-wise bidirectional latent embedding framework. When a mapping is done from high-dimensional feature space to a low-dimensional semantic space using regressors,  the distinctive features will partially fade while in the visual feature space, the structures are better preserved. Hence, the visual embedding space is well-known for mitigating the hubness problem. \cite{wan2019transductive} and \cite{zhang2017learning} use the output of the visual space of the CNN as the embedding space.\\

\textbf{Domain-shift.} Zero-shot learning challenge can be considered as a domain adaptation problem. This is because the source labelled data is disjoint with the target unlabelled domain data. This is called project domain-shift. Domain adaptation techniques are used to learn the intrinsic relationships among these domains and transfer knowledge between the two. A considerable amount of works has been done through a transductive setting which has been successful to overcome the domain-shift issue. \cite{fu2015transductive} a  multi-view embedding framework, performs label propagation on graph a heuristic one-stage self-learning approach to assign points to their nearest data points. \cite{kodirov2015unsupervised} introduces a regularised sparse coding based unsupervised domain adaptation framework that solves the domain shift problem. \cite{zhang2016zerods} use a structured prediction method to solve the problem by visually clustering the unseen data. \cite{wan2019transductive} use a visual constraint on the centre of each class when the mapping is being learned. Since the pure definition of the ZSL challenge is the inaccessibility of unseen data during training, several inductive approaches tried to solve the problem as well. \cite{kodirov2017semantic} propose to reconstruct the visual features to alleviate this issue. \cite{ye2017zero} perform sparse non-negative matrix factorisation for both domains in a common semantic dictionary. MFMR \cite{xu2017matrix} exploits the manifold structure of test data with a joint prediction scheme to avoid domain shift. \cite{rostami2019zero} use entropy minimisation in optimisation. \cite{li2019generalized} preserve the semantic similarity structure in seen and unseen classes to avoid the domain-shift occurrence. \cite{li2019leveraging} mitigates projection domain-shift by generating soul samples that are related to the semantic descriptions.

These three common issues together with inferiorities of each methods will be a motivation to decide on a particular approach when solving the ZSL problem. Attribute classifiers are considered customised since human-annotations are used; however, this makes the problem a laborious task that has strong supervision. Compatibility learning approaches have the ability to learn directly by eliminating the intermediate step but often face with the bias and hubness problem. Manifold learning solves this weakness of the semantic learning approaches by preserving the geometrical structure of the features. Cross-modal latent embedding approaches take on a different point of view and leverage both visual and semantic features and the similarity and differences between them. They often propose methods for aligning the structures between the two modes of features. This category of methods also suffers from the hubness problem for the problems dealing with high-dimensional data. Visual space embedding approaches have the advantage of turning the problem into a supervised one by generating or aggregating visual instances for the unseen classes. Plus are a favourable approach for solving hubness problem due to the 
high-dimensionality of the visual space that can preserve information structure better and also bias problem by alleviating the imbalanced data by generating unseen class samples. Here a challenge would be generating more realistic looking data. Another different setting is transductive learning that present solutions to bias problem, by creating balance in data by gathering unseen data, yet not applicable to many of the real-world problems since the original definition of ZSL limits the use of unseen data during the training phase.

Depending on the real-world scenarios, each way of solving the problem might be the most appropriate choice. Some approaches improve the solution by combining two or more methods to benefit from each one's strengths. 

\section{Conclusion \label{conclude}}
\label{conc}
\noindent In this article, we performed a comprehensive and multi-faceted review on the Zero-Shot/Generalised Zero-shot Learning challenge, its fundamentals, and variants for different scenarios and applications such as COVID-19 diagnosis, Autonomous Vehicles, and similar complex real-world applications which involve fully/partially new concepts that have never/rarely seen before, besides the barrier of limited annotated dataset. We divided the recent state-of-the-art methods into four space-wise embedding categories. We also reviewed different types of side and auxiliary information. We went through the popular datasets and their corresponding splits for the problem of ZSL. The paper also contributed in performing the experiment results for some of the common baselines and elaborated on assessing the advantages and disadvantages of each group, as well as the ideas behind different areas of solutions to improve each group.
Our evaluation reveals that data synthesis methods and combinational approaches yield the best performance, as by synthesising data, the problem shifts to the classic recognition/diagnosis problem, and by combining other methods, the model utilises the advantage of each embedding techniques. 
The models even outperform compatibility learning models in transductive setting. This means, the models consisting a visual data generation step, lead to better results than other approaches and settings. Furthermore, the accuracies improve when the unseen classes have closer semantic hierarchy and relatedness distance to the seen classes. 
Finally, we reviewed the current and potential real-world applications of ZSL and GZSL in the near future. To the best of our knowledge, such a comprehensive and detailed technical review and categorisation of the ZSL methodologies, alongside with an efficient solution for the recent challenge of COVID-19 pandemic is not done before; hence, we expect it to be helpful in developing new research directions among AI and health-related research community.

\bibliography{bibliography-sorted}

\end{document}